\documentclass[journal]{IEEEtran}
\usepackage{amsmath,amsfonts}
\usepackage{algorithmic}
\usepackage{algorithm}
\usepackage{array}
\usepackage[caption=false,font=footnotesize,labelfont=rm,textfont=rm]{subfig}

\usepackage{textcomp}
\usepackage{stfloats}
\usepackage{url}
\usepackage{verbatim}
\usepackage{graphicx}
\usepackage{cite}    
\makeatletter
\let\NAT@parse\undefined
\makeatother
\usepackage[colorlinks,linkcolor=blue,citecolor=blue]{hyperref}

\usepackage{booktabs}       
\usepackage{nicefrac}       
\usepackage{microtype}      
\usepackage{xcolor}         
\usepackage{multirow}       
\usepackage{bm}             
\usepackage{siunitx}
\newcommand{\rev}[1]{#1}

\begin{document}

\title{LightSAE: Parameter-Efficient and Heterogeneity-Aware Embedding for \\ IoT Multivariate Time Series Forecasting}

\author{Yi~Ren~and~Xinjie~Yu,~\IEEEmembership{Senior~Member,~IEEE}
\thanks{Y. Ren and X. Yu are with the Department of Electrical Engineering, Tsinghua University, Beijing 100084, China (e-mail: ren-y20@mails.tsinghua.edu.cn; yuxj@tsinghua.edu.cn).}
\thanks{Manuscript received [DATE]; revised [DATE].}}

\markboth{IEEE Internet of Things Journal,~Vol.~XX, No.~XX, [MONTH]~[YEAR]}%
{Ren \MakeLowercase{\textit{et al.}}: LightSAE: Parameter-Efficient and Heterogeneity-Aware Embedding for IoT Multivariate Time Series Forecasting}


\maketitle

\begin{abstract}
Modern Internet of Things (IoT) systems generate massive, heterogeneous multivariate time series data. Accurate Multivariate Time Series Forecasting (MTSF) of such data is critical for numerous applications. However, existing methods almost universally employ a shared embedding layer that processes all channels identically, creating a representational bottleneck that obscures valuable channel-specific information. To address this challenge, we introduce a Shared-Auxiliary Embedding (SAE) framework that decomposes the embedding into a shared base component capturing common patterns and channel-specific auxiliary components modeling unique deviations. Within this decomposition, we \rev{empirically observe} that the auxiliary components tend to exhibit low-rank and clustering characteristics, a structural pattern that is significantly less apparent when using purely independent embeddings.
Consequently, we design LightSAE, a parameter-efficient embedding module that operationalizes these observed characteristics through low-rank factorization and a shared, gated component pool. Extensive experiments across 9 IoT-related datasets and 4 backbone architectures demonstrate LightSAE's effectiveness, achieving MSE improvements of up to 22.8\% with only 4.0\% parameter increase. Code is available at https://github.com/EDM314/LightSAE.

\end{abstract}


\section{Introduction}
\label{sec:introduction}
\IEEEPARstart{T}{he} widespread adoption of Internet of Things (IoT) systems has generated massive volumes of multivariate time series data from diverse sensors and applications across various domains. Accurately forecasting such IoT-generated data has become crucial for optimizing system performance and enabling intelligent decision-making in applications such as energy load prediction and traffic flow management\cite{10980332,zhang2024sageformer}. This growing demand has driven substantial research advances in deep learning architectures for Multivariate Time Series Forecasting (MTSF), leading to significant improvements in prediction accuracy \cite{qiu2024tfb}.

\begin{figure}[t]
\centering
\subfloat[]{\includegraphics[width=.99\linewidth]{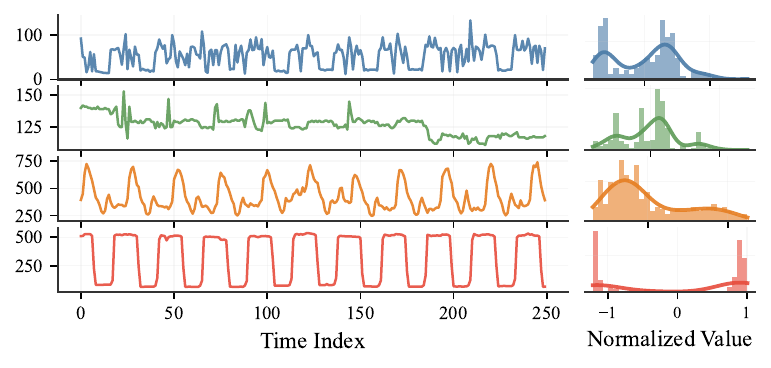}}   
\vspace{-1em}
\subfloat[]{\includegraphics[width=.9\linewidth]{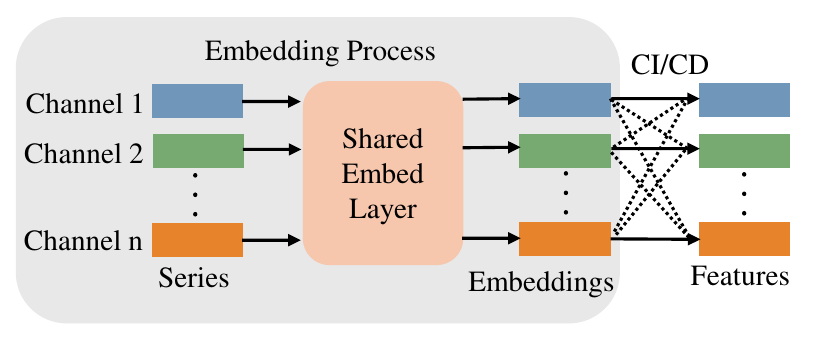}}
\caption{Illustration of our motivation. (a) Channel heterogeneity in the Electricity dataset~\cite{electricityloaddiagrams20112014_321}. Four representative channels demonstrate distinct temporal patterns and statistical distributions: irregular fluctuations (top), trend with periodic spikes (second), regular oscillations (third), and square-wave patterns (bottom). (b) Existing MTSF methods uniformly adopt a shared embedding layer.}
\label{fig:channel_heterogeneity}
\end{figure}

However, a significant challenge in IoT-driven MTSF is channel heterogeneity: different channels often originate from distinct sensor types, measure different physical phenomena, or exhibit unique temporal dynamics and statistical distributions \cite{kim2021reversible,liu2022non}. As illustrated in Fig.~\ref{fig:channel_heterogeneity}(a), channels can display dramatically different patterns, from irregular fluctuations to periodic oscillations, with corresponding variations in their underlying distributions. This heterogeneity suggests that channels may require specialized representational treatment to capture their unique characteristics effectively.

Despite this clear need for channel-specific modeling, existing MTSF methods have largely overlooked this challenge. Both channel-independent (CI)~\cite{nie2022time,li2023revisiting} and channel-dependent (CD)~\cite{liu2024itransformer} paradigms almost universally adopt a shared embedding layer (Fig.~\ref{fig:channel_heterogeneity}(b)), forcing heterogeneous channels through an identical transformation. \rev{This shared embedding strategy is conceptually analogous to using a single shared encoder to embed different data modalities (e.g., image and text). In fields like multimodal learning~\cite{DBLP:conf/icml/2021,zhang2023meta}, using modality-specific encoders~\cite{chen2025operational,team2025gemma} is a well-established and effective practice at preventing the loss of unique information and improving model performance. The underlying reason is that a shared encoder, optimized for an average case, can obscure modality-specific signals, creating an informational bottleneck. However, this design has been largely under-explored in MTSF for handling channel heterogeneity.}

This shared embedding limitation can be particularly impactful because it occurs at the initial stage of representation learning. When the initial embedding conflates channel-specific signals, valuable information risks being lost, resulting in compromised representations. Subsequent layers, regardless of their architectural sophistication, are limited to operating on these representations that lack channel-specific information. For instance, even advanced CD methods like iTransformer~\cite{liu2024itransformer}, which attempt to model intricate cross-channel interactions in deeper layers, are still constrained by the compromised channel representations produced by their shared embedding layers.

To address this challenge, we propose the Shared-Auxiliary Embedding (SAE) framework, which decomposes the embedding into a shared base component that captures global patterns common to all channels, and channel-specific auxiliary components that learn tailored representations for each channel's unique characteristics. However, directly implementing SAE faces a significant scalability challenge: when the number of channels is large, the parameter cost grows substantially due to the separate auxiliary components for each channel.

Fortunately, through deeper analysis of the SAE framework, we observe that the learned auxiliary component weights exhibit structural patterns that offer a pathway to address this scalability challenge: specifically, \textbf{low-rank and clustering characteristics}. The low-rank characteristic suggests that channel-specific deviations from common patterns can be compactly represented, while the clustering characteristic indicates that similar channels can share representational components. Notably, these characteristics are more clearly observed when auxiliary components are learned in conjunction with a shared base, and are weaker in purely independent channel embeddings without our SAE decomposition.

Building on this observation, we design LightSAE, a parameter-efficient embedding module that operationalizes these observed structures to achieve effective channel heterogeneity modeling with minimal parameter overhead. Our LightSAE addresses SAE's scalability challenge through two synergistic mechanisms: (1) low-rank factorization to compactly represent channel-specific deviations while preserving their expressiveness, and (2) a shared pool of components with a gating mechanism that facilitates both the reuse of common patterns among similar channels and the selection of specialized patterns for distinct ones, thereby leveraging the natural clustering of heterogeneous behaviors. The resulting LightSAE module is a parameter-efficient, plug-and-play component that effectively captures channel heterogeneity with substantially reduced parameter cost.

Our main contributions can be summarized as follows:
\begin{itemize}
    \item We introduce the Shared-Auxiliary Embedding (SAE) framework and, through its analysis, observe that channel-specific auxiliary components, when disentangled from common patterns, tend to exhibit low-rank and clustering characteristics.
    \item We design LightSAE, a parameter-efficient embedding module, which operationalizes these observed characteristics through low-rank factorization and a shared, gated component pool.
    \item We empirically validate LightSAE's effectiveness across 9 IoT-related datasets spanning diverse domains and 4 backbone architectures, achieving up to 22.8\% MSE improvement with only 4.0\% parameter increase.
    \item We conduct comprehensive ablation studies to dissect LightSAE's mechanisms, highlighting the role of the shared base in making these structures more apparent and the synergistic benefits of leveraging both low-rank and clustering observations. Furthermore, our experiments suggest that modeling channel heterogeneity at the initial embedding stage is more effective than applying it at later stages.
\end{itemize}

\section{Related Work}
\label{sec:related_work}
\rev{
\subsection{Deep Learning-based Time Series Forecasting}
Deep learning approaches for MTSF are broadly categorized into channel-independent (CI) and channel-dependent (CD) methods~\cite{wang2024deep}. CI models such as DLinear~\cite{zeng2023transformers}, RLinear~\cite{li2023revisiting}, and PatchTST~\cite{nie2022time} treat each channel as a separate univariate problem with shared parameters; CD models like iTransformer~\cite{liu2024itransformer}, TimesNet~\cite{wu2023timesnet}, and TSMixer~\cite{chen2023tsmixer} explicitly model cross-channel interactions. Despite architectural differences, both paradigms almost universally adopt a single shared embedding layer, creating an initial representational bottleneck that can obscure valuable channel-specific information. Our work directly targets this bottleneck by rethinking the embedding layer itself as a channel‑specific transformation.

\subsection{Early-Stage Heterogeneity Modeling}
Our focus on embedding-stage heterogeneity is motivated by a concept well-established in other domains. In fields like multimodal learning~\cite{DBLP:conf/icml/2021,zhang2023meta} and meteorological forecasting~\cite{chen2025operational,zhu2024puyun}, it is standard practice to use specialized encoders for heterogeneous data sources (e.g., image vs. text, or temperature vs. pressure) to preserve unique statistical characteristics and prevent information loss. This concept of early-stage, source-specific encoding, however, has been largely overlooked in the general MTSF literature, where different channels are typically treated uniformly by a shared embedding layer. Our work addresses this gap by analyzing this issue and proposing LightSAE, which brings this established design into MTSF in a parameter-efficient manner.

\subsection{Modeling Channel Heterogeneity}
We contrast three lines of work with our approach.

\subsubsection{Identity and Spatio-Temporal Embeddings}
Methods such as STID~\cite{shao2022spatial}, STAEformer~\cite{liu2023spatio}, and GAFormer~\cite{xiao2024gaformer} address identity indistinguishability by augmenting a shared series embedding with additive positional or identity-based terms. In contrast, LightSAE models heterogeneity in the transformation function itself by learning a unique mapping for each channel. Our design is explicitly driven by two structural observations we identify, low‑rankness and clusterability, to achieve parameter efficiency. These identity-based embeddings are complementary to our series‑transformation approach, not substitutes.

\subsubsection{Heterogeneity in Deep Layers}
Architectures like SageFormer~\cite{zhang2024sageformer} and iTransformer~\cite{liu2024itransformer} focus on modeling deep feature interactions across channels. These methods typically operate on representations generated by a shared embedding layer and employ sophisticated mechanisms (e.g., GNNs or attention across variates) within their deep encoder blocks to capture cross-channel dependencies. Their primary contribution lies in the interaction module, not in the initial representation learning of individual series. Our work addresses a different, upstream problem: we focus on modeling heterogeneity at the initial embedding stage itself, allowing downstream feature interaction modules to operate more effectively. The two directions are therefore highly complementary.

\subsubsection{Parameter-Efficient and Expert-Based Models}
C‑LoRA~\cite{nie2024channel} applies channel‑wise low‑rank adaptation on post‑embedding features, while MoE‑style approaches such as MoLE~\cite{ni2024mixture} and VE~\cite{wang2024ve} select among experts at the output layer. Compared to these, LightSAE differs in two primary aspects: (i) Architectural Positioning: LightSAE is applied at the initial embedding layer to preserve channel-specific information from the outset. In contrast, other methods act as late-stage modules, operating on feature representations that have already been processed by a shared embedding function. (ii) Structural Motivation: Our design is directly motivated by the structural patterns observed in our SAE analysis. We first observe that channel-specific auxiliary components exhibit distinct low-rank and clustering characteristics. LightSAE is then purpose-built to incorporate these observations as inductive biases. While other works employ similar techniques like LoRA or gating, they lack a similar structural analysis. Crucially, as our analysis shows, these structural observations are significantly weaker in purely independent embeddings and emerge more clearly under our SAE decomposition. This finding highlights the value of our integrated framework, as it creates a favorable structural context for these parameter-efficient techniques to be most effective.
}
\section{Revisiting Embedding Mechanisms for MTSF}
\label{sec:revisiting_embeddings}

This section revisits embedding mechanisms for MTSF, where each channel typically corresponds to data from an individual IoT sensor. For clarity, our primary focus is on series-level embedding strategies~\cite{liu2024itransformer,wang2024timexer,li2023revisiting}, where the entire time series of each individual channel is first mapped to a latent representation. The underlying ideas are conceptually extendable to patch-level approaches~\cite{nie2022time,zhang2023crossformer,liu2024unitst}.
\subsection{General Pipeline of MTSF Models}
\label{ssec:general_pipeline_mtsf}
As illustrated in Fig.~\ref{fig:framework}(a), a general deep learning model for MTSF adopting such an initial per-channel embedding approach first maps the input series $\bm{X} \in \mathbb{R}^{N \times L}$ to a set of channel-wise embeddings $\{\bm{e}_1, \dots, \bm{e}_N\}$, where $\bm{e}_i \in \mathbb{R}^{d_{\mathrm{model}}}$ is the embedding for the $i$-th channel's series $\bm{X}_{i,:} \in \mathbb{R}^{1 \times L}$ (denoted $\bm{X}_i$ for brevity). This can be expressed as:
\begin{equation}
\label{eq:general_pipeline_embedding}
\bm{e}_i = f_{\mathrm{emb}}(\bm{X}_i; \Theta_{\mathrm{emb}}) \quad \text{for } i=1, \dots, N
\end{equation}
These $N$ embeddings are then typically processed by a backbone network (e.g., Transformers~\cite{liu2024itransformer}, MLPs~\cite{li2023revisiting}) to capture complex dependencies and interactions, yielding a set of feature representations $\bm{H}_{\mathrm{feat}}$:
\begin{equation}
\label{eq:general_pipeline_backbone}
\bm{H}_{\mathrm{feat}} = \mathrm{Backbone}(\{\bm{e}_i\}_{i=1}^N; \Theta_{\mathrm{backbone}})
\end{equation}
Finally, a projection head maps these features to the desired forecast horizon $H$ for all $N$ channels:
\begin{equation}
\label{eq:general_pipeline_projection}
\bm{\hat{Y}} = \mathrm{Head}(\bm{H}_{\mathrm{feat}}; \Theta_{\mathrm{proj}}) \in \mathbb{R}^{N \times H}
\end{equation}

The entire model, comprising $\Theta_{\mathrm{emb}}$, $\Theta_{\mathrm{backbone}}$, and $\Theta_{\mathrm{proj}}$, is trained end-to-end by minimizing a loss function $\mathcal{L}$, typically the Mean Squared Error (MSE) between the prediction $\bm{\hat{Y}}$ and the ground truth $\bm{Y}$:
\begin{equation}
\label{eq:loss_mse}
\mathcal{L}(\bm{Y}, \bm{\hat{Y}}) = \frac{1}{N \cdot H} \sum_{i=1}^{N} \sum_{j=1}^{H} (Y_{i,j} - \hat{Y}_{i,j})^2.
\end{equation}

\subsection{The Standard Shared Embedding Strategy}
\label{ssec:shared_embedding_limitation}

\begin{figure*}[h]
    \centering
    \includegraphics[width=0.7\linewidth]{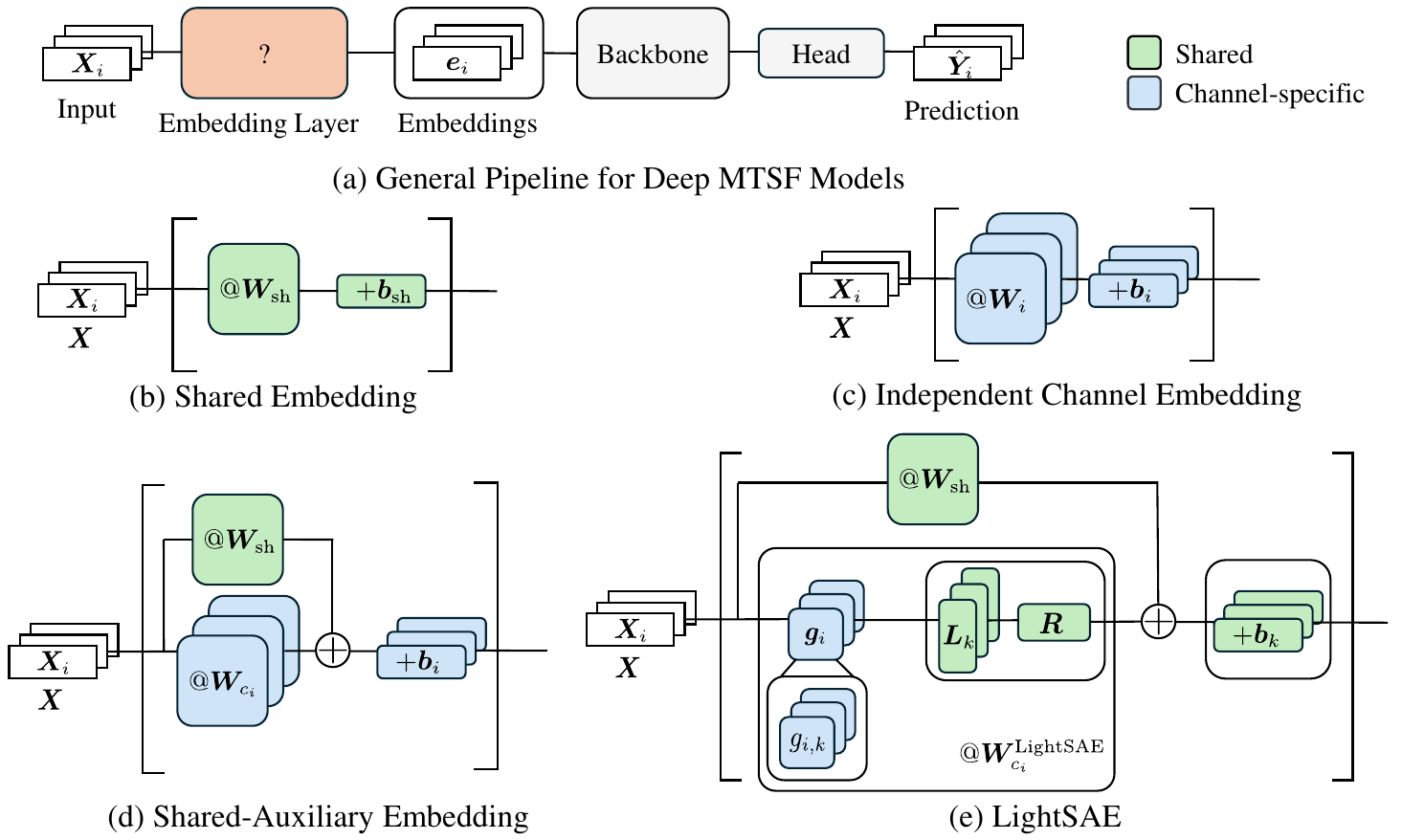}
    \caption{A comparison of different embedding strategies for MTSF. (a) The general pipeline for deep MTSF models. (b) Standard Shared Embedding, where all channels use a single embedding layer. (c) Independent Channel Embedding, where each channel has its own separate embedding layer. (d) Our proposed Shared-Auxiliary Embedding (SAE), which combines a shared base with channel-specific auxiliary components. (e) Our final LightSAE module, which enhances SAE with parameter-efficient low-rank components and a component pool.}
    \label{fig:framework}
\end{figure*}

Contemporary MTSF models typically employ a \textbf{Shared Embedding} approach, as shown in Fig.~\ref{fig:framework}(b), for the initial per-channel series-level embedding. Given an input $\bm{X} \in \mathbb{R}^{N \times L}$ with $N$ channels and lookback window length $L$, this approach uses a single shared linear transformation to embed each channel's series $\bm{X}_i \in \mathbb{R}^{1 \times L}$ into a $d_{\mathrm{model}}$-dimensional vector:

\begin{equation}
    \label{eq:shared_emb}
\bm{e}_i^{\mathrm{shared}} = \bm{X}_i \bm{W}_{\mathrm{sh}} + \bm{b}_{\mathrm{sh}},
\end{equation}
where $\bm{W}_{\mathrm{sh}} \in \mathbb{R}^{L \times d_{\mathrm{model}}}$ and $\bm{b}_{\mathrm{sh}} \in \mathbb{R}^{d_{\mathrm{model}}}$ are shared across all channels. For brevity, bias terms are omitted in subsequent equations unless explicitly stated. While this strategy is parameter-efficient, it overlooks significant channel heterogeneity and may obscure channel-specific information, thereby limiting model performance.

\subsection{Shared-Auxiliary Embedding (SAE) Architecture}
\label{ssec:sae_for_heterogeneity}
To learn channel-specific characteristics, a simple and intuitive approach is the \textbf{Independent Channel Embedding}, depicted in Fig.~\ref{fig:framework}(c), which uses a separate linear transformation for each channel:

\begin{equation}
\label{eq:ind_emb}
\bm{e}_i^{\mathrm{ind}} = \bm{X}_i \bm{W}_i,
\end{equation}
where $\bm{W}_i \in \mathbb{R}^{L \times d_{\mathrm{model}}}$ are the weights for channel $i$. While this approach offers flexibility, it suffers from two limitations: (1) high parameter costs ($N \times L \times d_{\mathrm{model}}$ total parameters) that can significantly exceed the backbone model size when $N$ is large, and (2) neglect of cross-channel commonalities, as each channel learns independently without leveraging shared patterns. These factors lead to high parameter overhead and overfitting risk, particularly in high-dimensional scenarios.

To address the challenge of channel heterogeneity while mitigating the limitations of both approaches above, we propose the \textbf{Shared-Auxiliary Embedding (SAE)} framework, shown in Fig.~\ref{fig:framework}(d). SAE balances the trade-off between expressiveness and parameter efficiency by decomposing the embedding into two complementary components: a shared base that captures cross-channel commonalities (addressing the limitation of Independent Channel Embedding) and channel-specific auxiliary components that model heterogeneous patterns (addressing the limitation of Shared Embedding):

\begin{equation}
\label{eq:sae_emb}
\bm{e}_i^{\mathrm{SAE}} = \underbrace{\bm{X}_i \bm{W}_{\mathrm{sh}}}_{\text{Shared Base}} + \underbrace{\bm{X}_i \bm{W}_{c_i}}_{\text{Auxiliary}},
\end{equation}
where $\bm{W}_{c_i} \in \mathbb{R}^{L \times d_{\mathrm{model}}}$ are auxiliary component weight parameters for channel $i$ (corresponding auxiliary biases $\bm{b}_{c_i}$ also exist).

While this formulation is more expressive than pure shared embedding, it still faces significant parameter scalability challenges when the number of channels is large. Fortunately, we observe that models trained with the SAE structure exhibit auxiliary weights $\bm{W}_{c_i}$ with structural characteristics, which can inform a more parameter-efficient design.

\subsection{Observations on Auxiliary Component Weights under SAE}
\label{ssec:aux_weight_properties_sae}

The decomposition provided by the SAE framework enables a focused analysis of the auxiliary weights $\bm{W}_{c_i}$. A key \rev{observation of our analysis} is that when disentangled from the common base component $\bm{W}_{\mathrm{sh}}$ within our SAE decomposition, the auxiliary weights $\bm{W}_{c_i}$ tend to exhibit two structural characteristics: \textbf{low-rankness} and \textbf{clusterability}. As we will show, these structures \rev{appear to be} weaker in the weights $\bm{W}_i$ from an Independent Channel Embedding, highlighting that these structural patterns become more apparent under our SAE decomposition. To the best of our knowledge, we are the first to report and leverage this structural observation in the context of MTSF.

\subsubsection{Low-Rank Characteristic of Auxiliary Component Weights}
\label{sssec:low_rank_property_sae}
In SAE,  $\bm{W}_{\mathrm{sh}}$ is designed to capture dominant common information, leaving $\bm{W}_{c_i}$ to model subtler, channel-specific adjustments.
We hypothesize that $\bm{W}_{c_i}$, representing deviations from the common patterns captured by $\bm{W}_{\mathrm{sh}}$, can be represented in a lower-dimensional space.

Performing Singular Value Decomposition (SVD) on learned $\bm{W}_{c_i} \in \mathbb{R}^{L \times d_{\mathrm{model}}}$ matrices from SAE, where $\bm{W}_{c_i} = \bm{U}_{c_i} \bm{\Sigma}_{c_i} \bm{V}_{c_i}^\top$ with $\bm{\Sigma}_{c_i} = \mathrm{diag}(\sigma_1, \sigma_2, \dots, \sigma_r)$ containing singular values $\sigma_j$ in descending order. We analyze the cumulative energy ratio $E_k$ of the first $k$ singular values:
\begin{equation}
\label{eq:cumulative_energy}
E_k = \frac{\sum_{j=1}^{k} \sigma_j^2}{\sum_{j=1}^{r} \sigma_j^2},
\end{equation}
where $r = \min(d_{\mathrm{model}}, L)$.

\begin{figure}[h]
    \centering
    {\footnotesize{Electricity}}\\
    \includegraphics[width=0.9\linewidth]{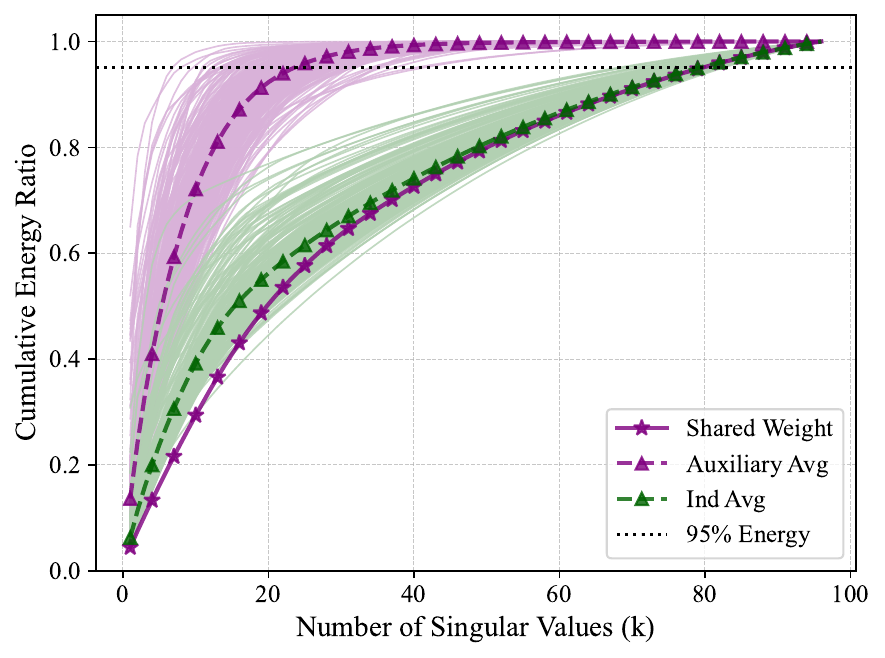} \\[-0.3em]
    {\footnotesize{PEMS04}}\\
    \includegraphics[width=0.9\linewidth]{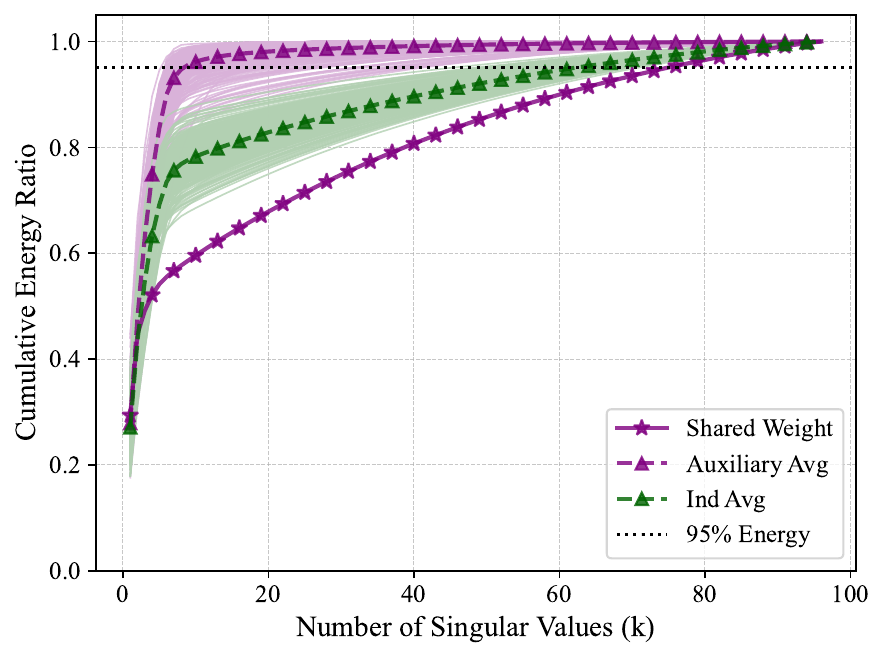}
    \caption{Cumulative energy ratio comparison for different embedding mechanisms. ``Shared Weight'' represents the shared component in SAE, ``Auxiliary Avg'' represents the averaged value of auxiliary components $\bm{W}_{c_i}$ from SAE, and ``Ind Avg'' represents the averaged value from Independent Channel Embedding weights, $\bm{W}_i$.
    }
    \label{fig:svd_decay_wc}
\end{figure}

The cumulative energy analysis generally shows that $\bm{W}_{c_i}$ matrices can rapidly achieve high energy ratios (e.g., 95\%) with a small number of leading components, as shown in Fig.~\ref{fig:svd_decay_wc} by the ``Auxiliary Avg'' curves. This indicates that $\bm{W}_{c_i}$ can be well-approximated by low-rank matrices.

In contrast, cumulative energy analysis on channel-specific weights $\bm{W}_i$ from models using Independent Channel Embedding often shows slower energy accumulation. The ``Ind Avg'' curves demonstrate this slower convergence. This suggests $\bm{W}_{\mathrm{sh}}$ in SAE isolates the common, higher-rank components, leaving $\bm{W}_{c_i}$ to represent lower-rank channel-specific deviations.

\subsubsection{Clustering Characteristic of Auxiliary Component Weights}
\label{sssec:clustering_property_sae}
We also find that auxiliary component weights $\bm{W}_{c_i}$ from SAE exhibit clustering behavior. Channels sharing similar deviation patterns from $\bm{W}_{\mathrm{sh}}$ may have similar $\bm{W}_{c_i}$, forming clusters.

\begin{figure}[h]
    \centering
    \subfloat[]{\includegraphics[width=0.49\linewidth]{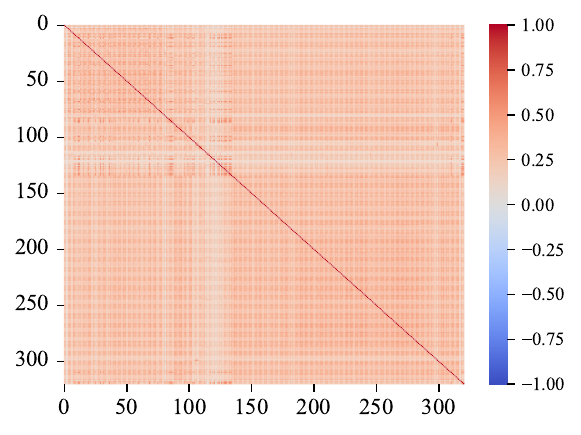}}
    \hspace{0em}
    \subfloat[]{\includegraphics[width=0.49\linewidth]{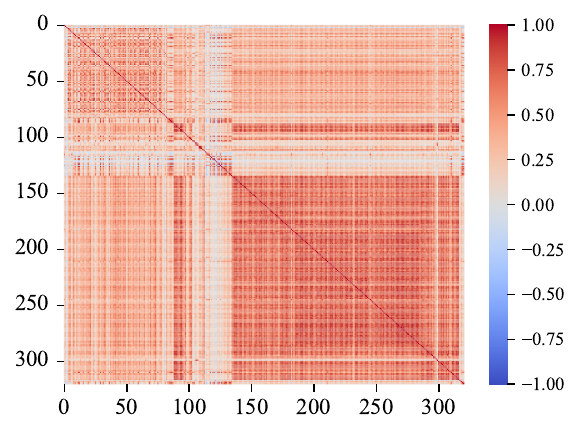}}
    \\ [-1em]
    \subfloat[]{\includegraphics[width=0.49\linewidth]{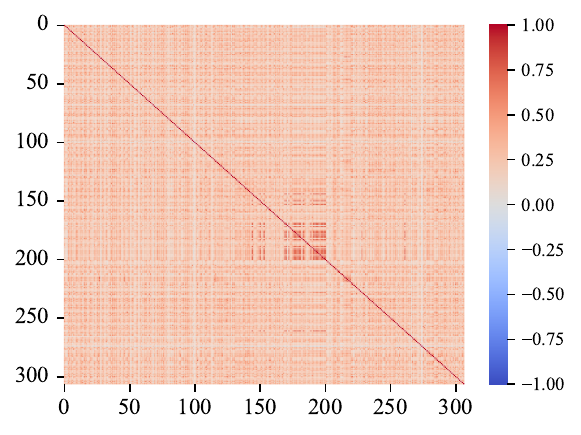}}
    \hspace{0em}
    \subfloat[]{\includegraphics[width=0.49\linewidth]{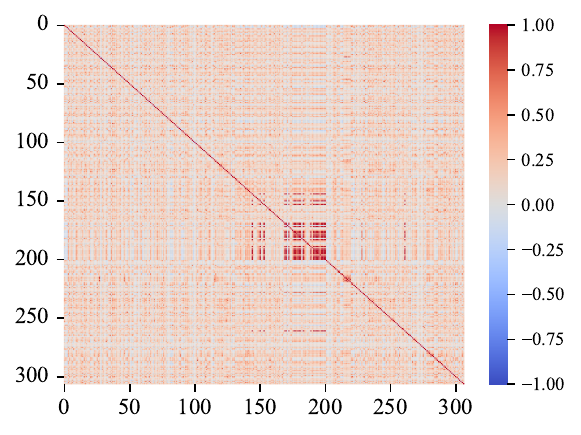}}
    \caption{Comparison of cosine similarity patterns between channel weights. (a) Similarity among Independent Channel Embedding weights $\bm{W}_i$ on Electricity dataset. (b) Similarity among SAE auxiliary weights $\bm{W}_{c_i}$ on Electricity dataset. (c) Similarity among Independent Channel Embedding weights $\bm{W}_i$ on PEMS04 dataset. (d) Similarity among SAE auxiliary weights $\bm{W}_{c_i}$ on PEMS04 dataset.}
    \label{fig:wc_clustering}
\end{figure}

In pairwise similarity analysis, the auxiliary weights $\bm{W}_{c_i}$ from SAE models exhibit distinct channel clusters when visualized via heatmaps. The cosine similarity between auxiliary weights of channels $p$ and $q$ is computed as:
\begin{equation}
\label{eq:cosine_similarity_wc}
\text{sim}(\bm{W}_{c_p}, \bm{W}_{c_q}) = \frac{\mathrm{vec}(\bm{W}_{c_p}) \cdot \mathrm{vec}(\bm{W}_{c_q})}{\|\mathrm{vec}(\bm{W}_{c_p})\|_2 \|\mathrm{vec}(\bm{W}_{c_q})\|_2},
\end{equation}
where $\mathrm{vec}(\cdot)$ denotes matrix vectorization. As demonstrated in Fig.~\ref{fig:wc_clustering}, SAE auxiliary weights exhibit sharp contrast with distinct high similarity values close to 1 and low similarity values near 0, forming clear block-diagonal clustering structures. In contrast, Independent Channel Embedding weights show more diffuse patterns with moderate similarity values distributed across a narrower range, resulting in less pronounced clustering boundaries. This clustering suggests shared types of heterogeneity captured by $\bm{W}_{c_i}$, and SAE's disentanglement of shared and auxiliary components makes these group structures more apparent.

\rev{
\subsubsection{Conceptual Explanation for the Observed Structures}
\label{sssec:gradient_intuition}
While a rigorous proof under realistic training dynamics is nontrivial and beyond our scope, we provide a concise, optimization-based intuition for why these structural characteristics emerge under the SAE framework. Let the per-channel losses be $\{\mathcal{L}_i\}_{i=1}^{N}$ and define the average loss $\mathcal{L} = \frac{1}{N}\sum_{i=1}^{N}\mathcal{L}_i$. The gradients for the shared component ($\bm{W}_{\mathrm{sh}}$) and an auxiliary component ($\bm{W}_{c_i}$) are:
\begin{align}
    \frac{\partial \mathcal{L}}{\partial \bm{W}_{\mathrm{sh}}} &= \frac{1}{N}\sum_{i=1}^{N}
    \frac{\partial \mathcal{L}_i}{\partial \bm{e}_i}\,
    \frac{\partial \bm{e}_i}{\partial \bm{W}_{\mathrm{sh}}}, \label{eq:grad_w_sh} \\
    \frac{\partial \mathcal{L}}{\partial \bm{W}_{c_i}} &= \frac{1}{N}\,
    \frac{\partial \mathcal{L}_i}{\partial \bm{e}_i}\,
    \frac{\partial \bm{e}_i}{\partial \bm{W}_{c_i}}. \label{eq:grad_w_ci}
\end{align}
This formulation recovers two special cases. When $\bm{W}_{c_i}=0$, the update reduces to the shared-only case (Eq.~\eqref{eq:shared_emb}); when $\bm{W}_{\mathrm{sh}}=0$, it reduces to the purely independent case (Eq.~\eqref{eq:ind_emb}).

\textbf{Rank perspective.}
The observed low rank of the auxiliary weights is consistent with the optimization dynamics induced by the SAE decomposition rather than a universal guarantee. The update for $\bm{W}_{\mathrm{sh}}$ averages gradients across channels, thereby emphasizing directions repeatedly useful across the entire dataset. Consequently, $\bm{W}_{\mathrm{sh}}$ is driven to encode complex, cross-channel structures, tending to result in a higher effective rank. In contrast, each $\bm{W}_{c_i}$ is driven only by the gradient of channel $i$ and tasked with fitting the residual relative to $\bm{W}_{\mathrm{sh}}$. This residual-fitting task is conceptually aligned with the concept of intrinsic dimensionality~\cite{DBLP:conf/acl/AghajanyanGZ20}, which posits that task-specific adaptations are inherently low-dimensional~\cite{hu2022lora}. This decomposition creates a structural context for the implicit bias of gradient-based optimizers to manifest. For this simpler task, the implicit bias of gradient-based optimizers towards low-rank solutions~\cite{gunasekar2017implicit, kim2025lora} can manifest more strongly, guiding the learning process to find compact solution for $\bm{W}_{c_i}$. Empirically, we observe that such residuals are captured by a low-dimensional transformation, yielding a lower effective rank in $\bm{W}_{c_i}$.

In the purely independent case, however, $\bm{W}_i$ must fulfill the roles of both $\bm{W}_{\mathrm{sh}}$ and $\bm{W}_{c_i}$ from the SAE framework. It needs to encode both the high-rank common patterns and the low-rank channel-specific residuals simultaneously. This entanglement of tasks results in a higher effective rank for the combined transformation. Furthermore, from an information-flow perspective, the embedding layer is the first point of data encoding. To preserve the high-rank information and maintain expressive power, the optimization process favors learning a high-rank transformation. Forcing a low-rank constraint here would risk a information bottleneck, as such a transformation is mathematically irreversible ($\mathrm{rank}(\bm{A}\bm{B}) \le \min\{\mathrm{rank}(\bm{A}), \mathrm{rank}(\bm{B})\}$). Accordingly, $\bm{W}_i$ typically exhibits a high effective rank, consistent with Fig.~\ref{fig:svd_decay_wc}.

\textbf{Clustering perspective.}
The clustering characteristic also emerges from the SAE structure. With a shared anchor $\bm{W}_{\mathrm{sh}}$ fixing a common basis, the auxiliary weights $\{\bm{W}_{c_i}\}$ represent residual deviations in the same coordinate system~\cite{flury1984common,liu2012robust}, making channels with similar deviations tend to exhibit high cosine similarity, producing distinct clusters (Fig.~\ref{fig:wc_clustering}b,d).

In the independent setting there is no anchor, and the learned bases $\{\bm{W}_i\}$ may differ by rotations, scalings or signs; together with the entanglement of common and specific factors, this misalignment dilutes pairwise similarity and weakens cluster structure (Fig.~\ref{fig:wc_clustering}a,c).

Notably, the explanation above is based on the assumption that learnable, high-rank common information exists across channels. In scenarios where such shared structure is absent, the rationale for the emergence of these low-rank and clustering patterns would be undermined.
}

\subsubsection{Summary of Empirical Observations under SAE}
Our empirical analysis of the SAE framework identifies two notable characteristics of the auxiliary component weights $\bm{W}_{c_i}$ when disentangled from the shared base: low-rank and clustering. These structural patterns emerge more clearly under the SAE decomposition and are weaker in purely independent channel embeddings.

\section{LightSAE: A Parameter-Efficient Heterogeneity-Aware Embedding Module}
\label{ssec:lightsae_design} 

Building on the observations from the SAE framework, our proposed LightSAE module, illustrated in Fig.~\ref{fig:framework}(e), achieves parameter efficiency while effectively modeling channel heterogeneity. It accomplishes this through two synergistic mechanisms: (i) low-rank factorization to compress auxiliary components, and (ii) a shared component pool with gating to leverage clustering patterns.

\subsection{Low-Rank Factorization of Auxiliary Components}
The observed low-rank characteristic of auxiliary weights $\bm{W}_{c_i}$ directly motivates their approximation via low-rank decomposition, similar to LoRA~\cite{hu2022lora}:
\begin{equation}
\label{eq:lora_wc_component}
\bm{W}_{c_i} \approx \bm{L}_{c_i} \bm{R}_{c_i},
\end{equation}
\rev{where $\bm{L}_{c_i} \in \mathbb{R}^{L \times r}$ and $\bm{R}_{c_i} \in \mathbb{R}^{r \times d_{\mathrm{model}}}$, with rank $r \ll \min(L, d_{\mathrm{model}})$.} This reduces parameters for each channel's auxiliary component from $L \cdot d_{\mathrm{model}}$ to $r(L + d_{\mathrm{model}})$. However, this direct application does not leverage the clustering observation that allows different channels to benefit from similar auxiliary patterns rather than learning entirely independent ones.

\subsection{Shared Pool with Gating Mechanism}
To exploit the clustering observation while maintaining parameter efficiency, we introduce a shared pool of low-rank components. Instead of independent matrices $\bm{L}_{c_i}, \bm{R}_{c_i}$ for each channel, LightSAE uses $K$ shared left matrices \rev{ $\{\bm{L}_k \in \mathbb{R}^{L \times r} \}_{k=1}^K$ and a single right matrix $\bm{R}_{\text{pool}} \in \mathbb{R}^{r \times d_{\mathrm{model}}}$}.

Each channel $i$ composes its auxiliary component by selecting from this shared pool via gating weights $g_{i,k}$, where $\sum_{k=1}^{K} g_{i,k}=1$. In our implementation, we use learnable channel-specific gates: each channel has $K$ learnable logits that are normalized via softmax to produce the gating weights. This allows channel $i$ to form its customized auxiliary component:
\begin{equation}
\label{eq:lightsae_wc_final} 
\bm{W}_{c_i}^{\text{LightSAE}} = \left(\sum_{k=1}^{K} g_{i,k} \bm{L}_k\right) \bm{R}_{\text{pool}}.
\end{equation}

This gating mechanism serves a dual role: it enables channels with similar residual characteristics to reuse a small set of shared components, improving parameter efficiency, while also allowing dissimilar channels to select different components, which mitigates cross‑cluster interference and yields a customized transformation per channel.

The final embedding for channel $i$ combines the shared base with the composed auxiliary component:
\begin{equation}
\label{eq:lightsae_final_embedding}
\bm{e}_i^{\text{LightSAE}} = \bm{X}_i (\bm{W}_{\mathrm{sh}} + \bm{W}_{c_i}^{\text{LightSAE}}).
\end{equation}

\textbf{Parameter Efficiency.} LightSAE's parameter cost is $K \cdot L \cdot r + r \cdot d_{\mathrm{model}} + N \cdot K$, comprising the shared left matrices, single right matrix, and gating parameters. Given that $r \ll \min(L, d_{\mathrm{model}})$ and $K \ll N$ in typical settings, this is significantly more efficient than the naive SAE approach ($N \cdot L \cdot d_{\mathrm{model}}$) when $N$ is large, while still capturing channel heterogeneity through the observed structural patterns.

Importantly, LightSAE introduces no computational overhead during inference. Similar to LoRA, the weights can be pre-computed and merged in two steps: first, the gated auxiliary weights are combined as $\bm{W}_{c_i}^{\text{aux}} = \left(\sum_{k=1}^{K} g_{i,k} \bm{L}_k\right) \bm{R}_{\text{pool}}$, then merged with the shared base to form the final embedding matrix $\bm{W}_i^{\text{final}} = \bm{W}_{\text{sh}} + \bm{W}_{c_i}^{\text{aux}}$ for each channel. This results in each channel having a single merged weight matrix, allowing the final embedding computation to maintain identical FLOPs to standard embedding layers.

\textbf{Plug-and-Play Integration.} LightSAE is designed to replace standard shared embedding layers in existing MTSF models. Its output preserves the expected format of channel-wise embeddings, enabling straightforward integration with various backbone architectures. The subsequent analysis demonstrates its effectiveness across diverse models and datasets.

\section{Experiments}
\label{sec:experiments}

To validate our proposed LightSAE, we conduct extensive experiments to evaluate four key aspects: (i) the impact of modeling channel heterogeneity on forecasting performance, (ii) the mechanisms underlying LightSAE's effectiveness and parameter efficiency, (iii) hyperparameter sensitivity and robustness across different configurations, and (iv) visualization analysis to understand the learned weights and gating mechanisms. 
\subsection{Experimental Setup}
\label{ssec:exp_setup}

\noindent\textbf{Datasets.} Our evaluation is conducted on 9 widely-used datasets for IoT-related MTSF. These datasets are representative and widely used in the literature~\cite{liu2024itransformer,wu2023timesnet,nie2022time} and evaluation framework TSLib~\cite{wang2024deep}, detailed in Table~\ref{tab:datasets}. These IoT datasets vary significantly in their characteristics, including the number of channels (7-883), temporal resolution (5 minutes to 1 hour), and application domains.

\noindent\textbf{Backbone Models.} We integrate LightSAE with four diverse forecasting architectures:
\begin{itemize}
    \item \textbf{RLinear}~\cite{li2023revisiting}: A simple yet effective linear model that directly maps input sequences to predictions.
    \item \textbf{RMLP}~\cite{li2023revisiting}: An MLP-based model with RevIN normalization that processes sequences through feedforward layers.
    \item \textbf{PatchTST}~\cite{nie2022time}: A Transformer model that segments time series into patches and applies self-attention.
    \item \textbf{iTransformer}~\cite{liu2024itransformer}: An inverted Transformer that treats channels as tokens for multivariate modeling.
\end{itemize}

\begin{table}[ht]
\centering
\caption{Detailed information about the IoT-related datasets used in our experiments.}
\label{tab:datasets}
\begin{tabular}{ccccc}
\toprule
\textbf{Dataset} & \textbf{Channels} & \textbf{Timesteps} & \textbf{Interval} & \textbf{Domain} \\
\midrule
ETTh1 & 7 & 14,400 & 1 hour & Electricity \\
ETTh2 & 7 & 14,400 & 1 hour & Electricity \\
ETTm1 & 7 & 57,600 & 15 mins & Electricity \\
ETTm2 & 7 & 57,600 & 15 mins & Electricity \\
Weather & 21 & 52,696 & 10 mins & Weather \\
Solar & 137 & 52,560 & 10 mins & Energy \\
Electricity & 321 & 26,304 & 1 hour & Electricity \\
PEMS04 & 307 & 16,992 & 5 mins & Transportation \\
PEMS07 & 883 & 28,224 & 5 mins & Transportation \\
\bottomrule
\end{tabular}
\end{table}

\rev{
\noindent\textbf{Implementation Details.} To ensure a fair and reproducible comparison, all experiments were conducted using TSLib~\cite{wang2024deep}, a widely-adopted library for MTSF evaluation~\cite{liu2024itransformer,gao2024units,zhou2021informer}. We strictly followed its standard protocols for data splitting and normalization. For all experiments, we used a fixed prediction horizon of $H=96$ and evaluated across four lookback windows: $\{96, 192, 336, 720\}$. For all baseline models, we adopted their default hyperparameters as provided in TSLib and their original publications~\cite{li2023revisiting,ni2024mixture,nie2024channel} without any additional tuning. For models integrated with LightSAE, we kept its own hyperparameters ($r=25$; $K \in \{3, 7, 10\}$ depending on the dataset) consistent across all backbones and only tuned the learning rate on the validation set from $\{1e-4, 5e-4, 1e-3, 5e-3, 1e-2\}$.
}

\subsection{Main Results}
\label{ssec:main_results}


\begin{table*}[h]
\centering
\caption{Performance of LightSAE across different backbone models on 9 datasets. The prediction horizon is fixed at $H=96$ and results are shown for different lookback window sizes $L \in \{96, 192, 336, 720\}$. \textit{Avg} denotes the average performance across all lookback windows. \textcolor{red}{Red} indicates improved performance and \textcolor[HTML]{5CAA82}{Green} indicates degraded performance.}
\label{tab:main_results_full}
\resizebox{\textwidth}{!}{%
\renewcommand{\arraystretch}{1.15}
\begin{tabular}{p{0.2cm}c|cc|cc|r@{\hspace{1pt}}r|cc|cc|r@{\hspace{1pt}}r|cc|cc|r@{\hspace{1pt}}r|cc|cc|r@{\hspace{1pt}}r}

\toprule[1pt]

\multicolumn{2}{c|}{Models} & \multicolumn{2}{c}{RLinear} & \multicolumn{2}{c}{w/ LightSAE} & \multicolumn{2}{c|}{Improve.} & \multicolumn{2}{c}{RMLP} & \multicolumn{2}{c}{w/ LightSAE} & \multicolumn{2}{c|}{Improve.} & \multicolumn{2}{c}{iTransformer} & \multicolumn{2}{c}{w/ LightSAE} & \multicolumn{2}{c|}{Improve.} & \multicolumn{2}{c}{PatchTST} & \multicolumn{2}{c}{w/ LightSAE} & \multicolumn{2}{c}{Improve.} \\

\cmidrule(l{2pt}r{2pt}){3-4}\cmidrule(l{2pt}r{2pt}){5-6}\cmidrule(l{2pt}r{2pt}){7-8}\cmidrule(l{2pt}r{2pt}){9-10}\cmidrule(l{2pt}r{2pt}){11-12}\cmidrule(l{2pt}r{2pt}){13-14}\cmidrule(l{2pt}r{2pt}){15-16}\cmidrule(l{2pt}r{2pt}){17-18}\cmidrule(l{2pt}r{2pt}){19-20}\cmidrule(l{2pt}r{2pt}){21-22}\cmidrule(l{2pt}r{2pt}){23-24}\cmidrule(l{2pt}r{2pt}){25-26}

\multicolumn{2}{c|}{Metric} & MSE & MAE & MSE & MAE & MSE & MAE & MSE & MAE & MSE & MAE & MSE & MAE & MSE & MAE & MSE & MAE & MSE & MAE & MSE & MAE & MSE & MAE & MSE & MAE \\

\midrule[1pt]
\multirow{5}{*}{\rotatebox[origin=c]{90}{ETTh1}}
& \multicolumn{1}{|c|}{96} & 0.386 & 0.395 & 0.381 & 0.389 & {\textcolor{red}{1.26\%}} & {\textcolor{red}{1.49\%}} & 0.405 & 0.413 & 0.384 & 0.395 & {\textcolor{red}{5.30\%}} & {\textcolor{red}{4.24\%}} & 0.386 & 0.405 & 0.381 & 0.399 & {\textcolor{red}{1.31\%}} & {\textcolor{red}{1.37\%}} & 0.414 & 0.419 & 0.381 & 0.396 & {\textcolor{red}{7.85\%}} & {\textcolor{red}{5.48\%}} \\
& \multicolumn{1}{|c|}{192} & 0.437 & 0.424 & 0.436 & 0.420 & {\textcolor{red}{0.22\%}} & {\textcolor{red}{0.85\%}} & 0.460 & 0.444 & 0.440 & 0.425 & {\textcolor{red}{4.31\%}} & {\textcolor{red}{4.20\%}} & 0.441 & 0.436 & 0.431 & 0.428 & {\textcolor{red}{2.27\%}} & {\textcolor{red}{1.93\%}} & 0.460 & 0.445 & 0.429 & 0.424 & {\textcolor{red}{6.78\%}} & {\textcolor{red}{4.61\%}} \\
& \multicolumn{1}{|c|}{336} & 0.479 & 0.446 & 0.478 & 0.442 & {\textcolor{red}{0.17\%}} & {\textcolor{red}{0.94\%}} & 0.505 & 0.466 & 0.483 & 0.446 & {\textcolor{red}{4.37\%}} & {\textcolor{red}{4.31\%}} & 0.487 & 0.458 & 0.484 & 0.454 & {\textcolor{red}{0.54\%}} & {\textcolor{red}{0.97\%}} & 0.501 & 0.466 & 0.471 & 0.448 & {\textcolor{red}{6.08\%}} & {\textcolor{red}{3.89\%}} \\
& \multicolumn{1}{|c|}{720} & 0.481 & 0.470 & 0.478 & 0.462 & {\textcolor{red}{0.55\%}} & {\textcolor{red}{1.67\%}} & 0.514 & 0.490 & 0.486 & 0.469 & {\textcolor{red}{5.44\%}} & {\textcolor{red}{4.36\%}} & 0.503 & 0.491 & 0.472 & 0.466 & {\textcolor{red}{6.14\%}} & {\textcolor{red}{5.12\%}} & 0.500 & 0.488 & 0.477 & 0.471 & {\textcolor{red}{4.55\%}} & {\textcolor{red}{3.43\%}} \\
\cmidrule(l{10pt}r{10pt}){2-26}
& \multicolumn{1}{|c|}{Avg} & 0.446 & 0.434 & 0.443 & 0.428 & {\textcolor{red}{0.55\%}} & {\textcolor{red}{1.24\%}} & 0.471 & 0.453 & 0.448 & 0.434 & {\textcolor{red}{4.86\%}} & {\textcolor{red}{4.28\%}} & 0.454 & 0.448 & 0.442 & 0.437 & {\textcolor{red}{2.56\%}} & {\textcolor{red}{2.35\%}} & 0.469 & 0.455 & 0.440 & 0.435 & {\textcolor{red}{6.32\%}} & {\textcolor{red}{4.35\%}} \\
\midrule
\multirow{5}{*}{\rotatebox[origin=c]{90}{ETTh2}}
& \multicolumn{1}{|c|}{96} & 0.288 & 0.338 & 0.288 & 0.337 & {\textcolor[HTML]{5CAA82}{-0.08\%}} & {\textcolor{red}{0.34\%}} & 0.291 & 0.342 & 0.293 & 0.343 & {\textcolor[HTML]{5CAA82}{-0.70\%}} & {\textcolor[HTML]{5CAA82}{-0.42\%}} & 0.297 & 0.349 & 0.300 & 0.348 & {\textcolor[HTML]{5CAA82}{-0.92\%}} & {\textcolor{red}{0.22\%}} & 0.302 & 0.348 & 0.294 & 0.345 & {\textcolor{red}{2.48\%}} & {\textcolor{red}{0.91\%}} \\
& \multicolumn{1}{|c|}{192} & 0.374 & 0.390 & 0.376 & 0.392 & {\textcolor[HTML]{5CAA82}{-0.54\%}} & {\textcolor[HTML]{5CAA82}{-0.59\%}} & 0.380 & 0.396 & 0.375 & 0.392 & {\textcolor{red}{1.36\%}} & {\textcolor{red}{1.01\%}} & 0.380 & 0.400 & 0.378 & 0.397 & {\textcolor{red}{0.59\%}} & {\textcolor{red}{0.77\%}} & 0.388 & 0.400 & 0.374 & 0.395 & {\textcolor{red}{3.49\%}} & {\textcolor{red}{1.36\%}} \\
& \multicolumn{1}{|c|}{336} & 0.415 & 0.426 & 0.418 & 0.429 & {\textcolor[HTML]{5CAA82}{-0.83\%}} & {\textcolor[HTML]{5CAA82}{-0.80\%}} & 0.419 & 0.428 & 0.416 & 0.430 & {\textcolor{red}{0.55\%}} & {\textcolor[HTML]{5CAA82}{-0.55\%}} & 0.428 & 0.432 & 0.415 & 0.429 & {\textcolor{red}{3.02\%}} & {\textcolor{red}{0.62\%}} & 0.426 & 0.433 & 0.417 & 0.428 & {\textcolor{red}{2.22\%}} & {\textcolor{red}{1.27\%}} \\
& \multicolumn{1}{|c|}{720} & 0.420 & 0.440 & 0.428 & 0.442 & {\textcolor[HTML]{5CAA82}{-1.95\%}} & {\textcolor[HTML]{5CAA82}{-0.35\%}} & 0.427 & 0.443 & 0.427 & 0.441 & {\textcolor{red}{0.12\%}} & {\textcolor{red}{0.35\%}} & 0.427 & 0.445 & 0.431 & 0.447 & {\textcolor[HTML]{5CAA82}{-0.96\%}} & {\textcolor[HTML]{5CAA82}{-0.51\%}} & 0.431 & 0.446 & 0.432 & 0.444 & {\textcolor[HTML]{5CAA82}{-0.23\%}} & {\textcolor{red}{0.41\%}} \\
\cmidrule(l{10pt}r{10pt}){2-26}
& \multicolumn{1}{|c|}{Avg} & 0.374 & 0.398 & 0.378 & 0.400 & {\textcolor[HTML]{5CAA82}{-0.85\%}} & {\textcolor[HTML]{5CAA82}{-0.35\%}} & 0.379 & 0.402 & 0.378 & 0.402 & {\textcolor{red}{0.33\%}} & {\textcolor{red}{0.10\%}} & 0.383 & 0.407 & 0.381 & 0.405 & {\textcolor{red}{0.43\%}} & {\textcolor{red}{0.28\%}} & 0.387 & 0.407 & 0.379 & 0.403 & {\textcolor{red}{1.99\%}} & {\textcolor{red}{0.99\%}} \\
\midrule
\multirow{5}{*}{\rotatebox[origin=c]{90}{ETTm1}}
& \multicolumn{1}{|c|}{96} & 0.355 & 0.376 & 0.339 & 0.367 & {\textcolor{red}{4.59\%}} & {\textcolor{red}{2.50\%}} & 0.337 & 0.374 & 0.315 & 0.354 & {\textcolor{red}{6.45\%}} & {\textcolor{red}{5.42\%}} & 0.334 & 0.376 & 0.334 & 0.369 & {\textcolor{red}{0.14\%}} & {\textcolor{red}{1.97\%}} & 0.329 & 0.367 & 0.321 & 0.361 & {\textcolor{red}{2.28\%}} & {\textcolor{red}{1.66\%}} \\
& \multicolumn{1}{|c|}{192} & 0.391 & 0.392 & 0.376 & 0.383 & {\textcolor{red}{3.72\%}} & {\textcolor{red}{2.39\%}} & 0.379 & 0.391 & 0.358 & 0.379 & {\textcolor{red}{5.41\%}} & {\textcolor{red}{3.15\%}} & 0.377 & 0.392 & 0.373 & 0.389 & {\textcolor{red}{1.10\%}} & {\textcolor{red}{0.80\%}} & 0.367 & 0.385 & 0.365 & 0.386 & {\textcolor{red}{0.61\%}} & {\textcolor[HTML]{5CAA82}{-0.31\%}} \\
& \multicolumn{1}{|c|}{336} & 0.424 & 0.415 & 0.408 & 0.403 & {\textcolor{red}{3.67\%}} & {\textcolor{red}{2.88\%}} & 0.412 & 0.412 & 0.390 & 0.401 & {\textcolor{red}{5.45\%}} & {\textcolor{red}{2.58\%}} & 0.426 & 0.415 & 0.404 & 0.409 & {\textcolor{red}{5.23\%}} & {\textcolor{red}{1.43\%}} & 0.399 & 0.410 & 0.390 & 0.403 & {\textcolor{red}{2.16\%}} & {\textcolor{red}{1.71\%}} \\
& \multicolumn{1}{|c|}{720} & 0.487 & 0.450 & 0.473 & 0.438 & {\textcolor{red}{2.85\%}} & {\textcolor{red}{2.78\%}} & 0.478 & 0.447 & 0.455 & 0.439 & {\textcolor{red}{4.87\%}} & {\textcolor{red}{1.69\%}} & 0.491 & 0.459 & 0.479 & 0.451 & {\textcolor{red}{2.44\%}} & {\textcolor{red}{1.69\%}} & 0.454 & 0.439 & 0.460 & 0.449 & {\textcolor[HTML]{5CAA82}{-1.29\%}} & {\textcolor[HTML]{5CAA82}{-2.37\%}} \\
\cmidrule(l{10pt}r{10pt}){2-26}
& \multicolumn{1}{|c|}{Avg} & 0.414 & 0.408 & 0.399 & 0.397 & {\textcolor{red}{3.71\%}} & {\textcolor{red}{2.64\%}} & 0.401 & 0.406 & 0.380 & 0.393 & {\textcolor{red}{5.55\%}} & {\textcolor{red}{3.21\%}} & 0.407 & 0.411 & 0.397 & 0.404 & {\textcolor{red}{2.23\%}} & {\textcolor{red}{1.47\%}} & 0.387 & 0.400 & 0.384 & 0.400 & {\textcolor{red}{0.94\%}} & {\textcolor{red}{0.17\%}} \\
\midrule
\multirow{5}{*}{\rotatebox[origin=c]{90}{ETTm2}}
& \multicolumn{1}{|c|}{96} & 0.182 & 0.265 & 0.177 & 0.257 & {\textcolor{red}{2.80\%}} & {\textcolor{red}{2.87\%}} & 0.180 & 0.262 & 0.176 & 0.257 & {\textcolor{red}{2.37\%}} & {\textcolor{red}{2.16\%}} & 0.180 & 0.264 & 0.178 & 0.261 & {\textcolor{red}{1.24\%}} & {\textcolor{red}{1.01\%}} & 0.175 & 0.259 & 0.175 & 0.257 & {\textcolor{red}{0.22\%}} & {\textcolor{red}{0.80\%}} \\
& \multicolumn{1}{|c|}{192} & 0.246 & 0.304 & 0.242 & 0.299 & {\textcolor{red}{1.50\%}} & {\textcolor{red}{1.58\%}} & 0.246 & 0.303 & 0.237 & 0.297 & {\textcolor{red}{3.28\%}} & {\textcolor{red}{2.00\%}} & 0.250 & 0.309 & 0.249 & 0.308 & {\textcolor{red}{0.45\%}} & {\textcolor{red}{0.18\%}} & 0.241 & 0.302 & 0.240 & 0.300 & {\textcolor{red}{0.58\%}} & {\textcolor{red}{0.56\%}} \\
& \multicolumn{1}{|c|}{336} & 0.307 & 0.342 & 0.304 & 0.339 & {\textcolor{red}{0.93\%}} & {\textcolor{red}{0.81\%}} & 0.308 & 0.343 & 0.293 & 0.334 & {\textcolor{red}{4.72\%}} & {\textcolor{red}{2.38\%}} & 0.311 & 0.348 & 0.316 & 0.349 & {\textcolor[HTML]{5CAA82}{-1.65\%}} & {\textcolor[HTML]{5CAA82}{-0.39\%}} & 0.305 & 0.343 & 0.296 & 0.336 & {\textcolor{red}{2.88\%}} & {\textcolor{red}{1.90\%}} \\
& \multicolumn{1}{|c|}{720} & 0.407 & 0.398 & 0.404 & 0.396 & {\textcolor{red}{0.69\%}} & {\textcolor{red}{0.53\%}} & 0.407 & 0.398 & 0.390 & 0.391 & {\textcolor{red}{4.37\%}} & {\textcolor{red}{1.63\%}} & 0.412 & 0.407 & 0.419 & 0.408 & {\textcolor[HTML]{5CAA82}{-1.62\%}} & {\textcolor[HTML]{5CAA82}{-0.36\%}} & 0.402 & 0.400 & 0.397 & 0.396 & {\textcolor{red}{1.26\%}} & {\textcolor{red}{1.06\%}} \\
\cmidrule(l{10pt}r{10pt}){2-26}
& \multicolumn{1}{|c|}{Avg} & 0.285 & 0.327 & 0.282 & 0.323 & {\textcolor{red}{1.48\%}} & {\textcolor{red}{1.45\%}} & 0.285 & 0.327 & 0.274 & 0.320 & {\textcolor{red}{3.68\%}} & {\textcolor{red}{2.04\%}} & 0.288 & 0.332 & 0.290 & 0.332 & {\textcolor[HTML]{5CAA82}{-0.39\%}} & {\textcolor{red}{0.11\%}} & 0.281 & 0.326 & 0.277 & 0.322 & {\textcolor{red}{1.23\%}} & {\textcolor{red}{1.08\%}} \\
\midrule
\multirow{5}{*}{\rotatebox[origin=c]{90}{Weather}}
& \multicolumn{1}{|c|}{96} & 0.192 & 0.232 & 0.162 & 0.206 & {\textcolor{red}{15.49\%}} & {\textcolor{red}{11.41\%}} & 0.196 & 0.235 & 0.161 & 0.207 & {\textcolor{red}{17.61\%}} & {\textcolor{red}{11.92\%}} & 0.174 & 0.214 & 0.158 & 0.203 & {\textcolor{red}{9.19\%}} & {\textcolor{red}{5.28\%}} & 0.177 & 0.218 & 0.161 & 0.209 & {\textcolor{red}{8.79\%}} & {\textcolor{red}{4.14\%}} \\
& \multicolumn{1}{|c|}{192} & 0.240 & 0.271 & 0.208 & 0.247 & {\textcolor{red}{13.20\%}} & {\textcolor{red}{8.91\%}} & 0.240 & 0.271 & 0.207 & 0.249 & {\textcolor{red}{13.56\%}} & {\textcolor{red}{8.09\%}} & 0.221 & 0.254 & 0.208 & 0.249 & {\textcolor{red}{5.88\%}} & {\textcolor{red}{1.98\%}} & 0.225 & 0.259 & 0.208 & 0.251 & {\textcolor{red}{7.45\%}} & {\textcolor{red}{3.02\%}} \\
& \multicolumn{1}{|c|}{336} & 0.292 & 0.307 & 0.264 & 0.288 & {\textcolor{red}{9.61\%}} & {\textcolor{red}{6.23\%}} & 0.291 & 0.307 & 0.264 & 0.291 & {\textcolor{red}{9.42\%}} & {\textcolor{red}{5.30\%}} & 0.278 & 0.296 & 0.264 & 0.291 & {\textcolor{red}{5.17\%}} & {\textcolor{red}{1.61\%}} & 0.278 & 0.297 & 0.264 & 0.291 & {\textcolor{red}{5.16\%}} & {\textcolor{red}{1.94\%}} \\
& \multicolumn{1}{|c|}{720} & 0.364 & 0.353 & 0.343 & 0.338 & {\textcolor{red}{5.89\%}} & {\textcolor{red}{4.19\%}} & 0.363 & 0.353 & 0.343 & 0.343 & {\textcolor{red}{5.40\%}} & {\textcolor{red}{2.77\%}} & 0.358 & 0.347 & 0.354 & 0.350 & {\textcolor{red}{1.25\%}} & {\textcolor[HTML]{5CAA82}{-0.90\%}} & 0.354 & 0.348 & 0.344 & 0.343 & {\textcolor{red}{2.93\%}} & {\textcolor{red}{1.42\%}} \\
\cmidrule(l{10pt}r{10pt}){2-26}
& \multicolumn{1}{|c|}{Avg} & 0.272 & 0.291 & 0.244 & 0.270 & {\textcolor{red}{11.05\%}} & {\textcolor{red}{7.69\%}} & 0.272 & 0.291 & 0.244 & 0.273 & {\textcolor{red}{11.50\%}} & {\textcolor{red}{7.02\%}} & 0.258 & 0.278 & 0.246 & 0.273 & {\textcolor{red}{5.37\%}} & {\textcolor{red}{1.99\%}} & 0.259 & 0.280 & 0.244 & 0.274 & {\textcolor{red}{6.08\%}} & {\textcolor{red}{2.63\%}} \\
\midrule
\multirow{5}{*}{\rotatebox[origin=c]{90}{Solar}}
& \multicolumn{1}{|c|}{96} & 0.322 & 0.339 & 0.308 & 0.334 & {\textcolor{red}{4.28\%}} & {\textcolor{red}{1.41\%}} & 0.233 & 0.296 & 0.214 & 0.268 & {\textcolor{red}{8.33\%}} & {\textcolor{red}{9.34\%}} & 0.203 & 0.237 & 0.177 & 0.237 & {\textcolor{red}{12.85\%}} & {\textcolor{red}{0.18\%}} & 0.234 & 0.286 & 0.206 & 0.265 & {\textcolor{red}{11.79\%}} & {\textcolor{red}{7.31\%}} \\
& \multicolumn{1}{|c|}{192} & 0.359 & 0.356 & 0.347 & 0.354 & {\textcolor{red}{3.23\%}} & {\textcolor{red}{0.53\%}} & 0.260 & 0.316 & 0.236 & 0.284 & {\textcolor{red}{9.42\%}} & {\textcolor{red}{10.20\%}} & 0.233 & 0.261 & 0.213 & 0.259 & {\textcolor{red}{8.55\%}} & {\textcolor{red}{0.93\%}} & 0.267 & 0.310 & 0.223 & 0.268 & {\textcolor{red}{16.56\%}} & {\textcolor{red}{13.67\%}} \\
& \multicolumn{1}{|c|}{336} & 0.397 & 0.369 & 0.389 & 0.369 & {\textcolor{red}{2.12\%}} & {\textcolor{red}{0.07\%}} & 0.276 & 0.323 & 0.247 & 0.289 & {\textcolor{red}{10.45\%}} & {\textcolor{red}{10.39\%}} & 0.248 & 0.273 & 0.201 & 0.274 & {\textcolor{red}{18.77\%}} & {\textcolor[HTML]{5CAA82}{-0.24\%}} & 0.290 & 0.315 & 0.224 & 0.271 & {\textcolor{red}{22.72\%}} & {\textcolor{red}{13.97\%}} \\
& \multicolumn{1}{|c|}{720} & 0.397 & 0.356 & 0.391 & 0.362 & {\textcolor{red}{1.47\%}} & {\textcolor[HTML]{5CAA82}{-1.80\%}} & 0.273 & 0.316 & 0.241 & 0.281 & {\textcolor{red}{11.74\%}} & {\textcolor{red}{11.11\%}} & 0.249 & 0.275 & 0.207 & 0.272 & {\textcolor{red}{16.81\%}} & {\textcolor{red}{1.07\%}} & 0.289 & 0.317 & 0.223 & 0.272 & {\textcolor{red}{22.70\%}} & {\textcolor{red}{14.11\%}} \\
\cmidrule(l{10pt}r{10pt}){2-26}
& \multicolumn{1}{|c|}{Avg} & 0.369 & 0.355 & 0.359 & 0.355 & {\textcolor{red}{2.78\%}} & {\textcolor{red}{0.05\%}} & 0.261 & 0.313 & 0.234 & 0.281 & {\textcolor{red}{9.98\%}} & {\textcolor{red}{10.26\%}} & 0.233 & 0.262 & 0.200 & 0.260 & {\textcolor{red}{14.24\%}} & {\textcolor{red}{0.48\%}} & 0.270 & 0.307 & 0.219 & 0.269 & {\textcolor{red}{18.45\%}} & {\textcolor{red}{12.26\%}} \\
\midrule
\multirow{5}{*}{\rotatebox[origin=c]{90}{Electricity}}
& \multicolumn{1}{|c|}{96} & 0.201 & 0.281 & 0.187 & 0.266 & {\textcolor{red}{7.05\%}} & {\textcolor{red}{5.34\%}} & 0.190 & 0.272 & 0.155 & 0.246 & {\textcolor{red}{18.77\%}} & {\textcolor{red}{9.75\%}} & 0.148 & 0.240 & 0.134 & 0.231 & {\textcolor{red}{9.17\%}} & {\textcolor{red}{3.82\%}} & 0.181 & 0.270 & 0.149 & 0.239 & {\textcolor{red}{17.77\%}} & {\textcolor{red}{11.32\%}} \\
& \multicolumn{1}{|c|}{192} & 0.201 & 0.283 & 0.187 & 0.269 & {\textcolor{red}{6.73\%}} & {\textcolor{red}{4.94\%}} & 0.195 & 0.279 & 0.167 & 0.256 & {\textcolor{red}{14.49\%}} & {\textcolor{red}{8.21\%}} & 0.162 & 0.253 & 0.155 & 0.250 & {\textcolor{red}{4.49\%}} & {\textcolor{red}{1.07\%}} & 0.188 & 0.274 & 0.162 & 0.254 & {\textcolor{red}{13.92\%}} & {\textcolor{red}{7.46\%}} \\
& \multicolumn{1}{|c|}{336} & 0.215 & 0.298 & 0.203 & 0.284 & {\textcolor{red}{5.64\%}} & {\textcolor{red}{4.59\%}} & 0.212 & 0.296 & 0.184 & 0.274 & {\textcolor{red}{13.04\%}} & {\textcolor{red}{7.61\%}} & 0.178 & 0.269 & 0.168 & 0.265 & {\textcolor{red}{5.43\%}} & {\textcolor{red}{1.44\%}} & 0.204 & 0.293 & 0.180 & 0.274 & {\textcolor{red}{11.99\%}} & {\textcolor{red}{6.50\%}} \\
& \multicolumn{1}{|c|}{720} & 0.257 & 0.331 & 0.244 & 0.318 & {\textcolor{red}{5.02\%}} & {\textcolor{red}{4.06\%}} & 0.255 & 0.331 & 0.223 & 0.308 & {\textcolor{red}{12.61\%}} & {\textcolor{red}{6.97\%}} & 0.225 & 0.317 & 0.190 & 0.286 & {\textcolor{red}{15.68\%}} & {\textcolor{red}{9.77\%}} & 0.246 & 0.324 & 0.220 & 0.310 & {\textcolor{red}{10.73\%}} & {\textcolor{red}{4.41\%}} \\
\cmidrule(l{10pt}r{10pt}){2-26}
& \multicolumn{1}{|c|}{Avg} & 0.218 & 0.298 & 0.205 & 0.284 & {\textcolor{red}{6.11\%}} & {\textcolor{red}{4.73\%}} & 0.213 & 0.295 & 0.182 & 0.271 & {\textcolor{red}{14.73\%}} & {\textcolor{red}{8.13\%}} & 0.178 & 0.270 & 0.162 & 0.258 & {\textcolor{red}{8.69\%}} & {\textcolor{red}{4.02\%}} & 0.205 & 0.290 & 0.177 & 0.269 & {\textcolor{red}{13.61\%}} & {\textcolor{red}{7.42\%}} \\
\midrule
\multirow{5}{*}{\rotatebox[origin=c]{90}{PEMS04}}
& \multicolumn{1}{|c|}{96} & 1.127 & 0.812 & 1.012 & 0.758 & {\textcolor{red}{10.19\%}} & {\textcolor{red}{6.63\%}} & 0.262 & 0.358 & 0.192 & 0.298 & {\textcolor{red}{26.47\%}} & {\textcolor{red}{16.70\%}} & 0.150 & 0.262 & 0.115 & 0.225 & {\textcolor{red}{23.48\%}} & {\textcolor{red}{14.26\%}} & 0.291 & 0.389 & 0.166 & 0.269 & {\textcolor{red}{42.93\%}} & {\textcolor{red}{30.77\%}} \\
& \multicolumn{1}{|c|}{192} & 1.150 & 0.808 & 1.038 & 0.757 & {\textcolor{red}{9.77\%}} & {\textcolor{red}{6.37\%}} & 0.299 & 0.384 & 0.234 & 0.329 & {\textcolor{red}{21.66\%}} & {\textcolor{red}{14.29\%}} & 0.195 & 0.298 & 0.140 & 0.247 & {\textcolor{red}{28.24\%}} & {\textcolor{red}{17.32\%}} & 0.303 & 0.396 & 0.203 & 0.298 & {\textcolor{red}{32.91\%}} & {\textcolor{red}{24.75\%}} \\
& \multicolumn{1}{|c|}{336} & 0.838 & 0.645 & 0.764 & 0.610 & {\textcolor{red}{8.84\%}} & {\textcolor{red}{5.40\%}} & 0.276 & 0.363 & 0.231 & 0.327 & {\textcolor{red}{16.41\%}} & {\textcolor{red}{10.08\%}} & 0.192 & 0.294 & 0.167 & 0.265 & {\textcolor{red}{13.11\%}} & {\textcolor{red}{10.05\%}} & 0.284 & 0.379 & 0.208 & 0.302 & {\textcolor{red}{26.87\%}} & {\textcolor{red}{20.26\%}} \\
& \multicolumn{1}{|c|}{720} & 0.977 & 0.725 & 0.890 & 0.684 & {\textcolor{red}{8.97\%}} & {\textcolor{red}{5.68\%}} & 0.324 & 0.399 & 0.272 & 0.362 & {\textcolor{red}{15.88\%}} & {\textcolor{red}{9.24\%}} & 0.256 & 0.341 & 0.190 & 0.292 & {\textcolor{red}{25.64\%}} & {\textcolor{red}{14.47\%}} & 0.324 & 0.406 & 0.236 & 0.328 & {\textcolor{red}{27.28\%}} & {\textcolor{red}{19.28\%}} \\
\cmidrule(l{10pt}r{10pt}){2-26}
& \multicolumn{1}{|c|}{Avg} & 1.023 & 0.748 & 0.926 & 0.702 & {\textcolor{red}{9.44\%}} & {\textcolor{red}{6.02\%}} & 0.290 & 0.376 & 0.232 & 0.329 & {\textcolor{red}{20.10\%}} & {\textcolor{red}{12.58\%}} & 0.198 & 0.299 & 0.153 & 0.257 & {\textcolor{red}{22.62\%}} & {\textcolor{red}{14.02\%}} & 0.301 & 0.392 & 0.203 & 0.299 & {\textcolor{red}{32.50\%}} & {\textcolor{red}{23.76\%}} \\
\midrule
\multirow{5}{*}{\rotatebox[origin=c]{90}{PEMS07}}
& \multicolumn{1}{|c|}{96} & 1.096 & 0.795 & 1.001 & 0.741 & {\textcolor{red}{8.66\%}} & {\textcolor{red}{6.84\%}} & 0.306 & 0.378 & 0.236 & 0.314 & {\textcolor{red}{22.82\%}} & {\textcolor{red}{16.75\%}} & 0.139 & 0.245 & 0.105 & 0.204 & {\textcolor{red}{24.38\%}} & {\textcolor{red}{16.93\%}} & 0.346 & 0.404 & 0.195 & 0.277 & {\textcolor{red}{43.56\%}} & {\textcolor{red}{31.39\%}} \\
& \multicolumn{1}{|c|}{192} & 1.149 & 0.798 & 1.054 & 0.744 & {\textcolor{red}{8.29\%}} & {\textcolor{red}{6.83\%}} & 0.368 & 0.416 & 0.284 & 0.350 & {\textcolor{red}{22.80\%}} & {\textcolor{red}{15.79\%}} & 0.192 & 0.287 & 0.143 & 0.232 & {\textcolor{red}{25.43\%}} & {\textcolor{red}{19.28\%}} & 0.352 & 0.409 & 0.247 & 0.306 & {\textcolor{red}{29.98\%}} & {\textcolor{red}{25.15\%}} \\
& \multicolumn{1}{|c|}{336} & 0.821 & 0.631 & 0.758 & 0.596 & {\textcolor{red}{7.63\%}} & {\textcolor{red}{5.57\%}} & 0.325 & 0.385 & 0.272 & 0.341 & {\textcolor{red}{16.36\%}} & {\textcolor{red}{11.49\%}} & 0.191 & 0.283 & 0.156 & 0.240 & {\textcolor{red}{18.14\%}} & {\textcolor{red}{14.98\%}} & 0.322 & 0.388 & 0.230 & 0.305 & {\textcolor{red}{28.71\%}} & {\textcolor{red}{21.29\%}} \\
& \multicolumn{1}{|c|}{720} & 0.972 & 0.710 & 0.900 & 0.668 & {\textcolor{red}{7.40\%}} & {\textcolor{red}{5.88\%}} & 0.382 & 0.422 & 0.323 & 0.378 & {\textcolor{red}{15.35\%}} & {\textcolor{red}{10.39\%}} & 0.256 & 0.331 & 0.191 & 0.272 & {\textcolor{red}{25.52\%}} & {\textcolor{red}{17.78\%}} & 0.379 & 0.423 & 0.288 & 0.341 & {\textcolor{red}{23.92\%}} & {\textcolor{red}{19.54\%}} \\
\cmidrule(l{10pt}r{10pt}){2-26}
& \multicolumn{1}{|c|}{Avg} & 1.009 & 0.733 & 0.928 & 0.687 & {\textcolor{red}{8.00\%}} & {\textcolor{red}{6.28\%}} & 0.345 & 0.400 & 0.279 & 0.346 & {\textcolor{red}{19.33\%}} & {\textcolor{red}{13.61\%}} & 0.194 & 0.286 & 0.149 & 0.237 & {\textcolor{red}{23.37\%}} & {\textcolor{red}{17.24\%}} & 0.350 & 0.406 & 0.240 & 0.307 & {\textcolor{red}{31.54\%}} & {\textcolor{red}{24.34\%}} \\
\midrule
\multicolumn{2}{c|}{Dataset Avg} & \multicolumn{4}{c|}{} & \textcolor{red}{4.70\%} & \textcolor{red}{3.30\%} & \multicolumn{4}{c|}{} & \textcolor{red}{10.01\%} & \textcolor{red}{6.80\%} & \multicolumn{4}{c|}{} & \textcolor{red}{8.79\%} & \textcolor{red}{4.66\%} & \multicolumn{4}{c|}{} & \textcolor{red}{12.52\%} & \textcolor{red}{8.56\%} \\
\bottomrule[1pt]
\end{tabular}%
}
\end{table*}

Table~\ref{tab:main_results_full} presents comprehensive forecasting results across 9 datasets and 4 backbone architectures. LightSAE achieves notable performance gains across most settings, delivering average MSE improvements of 4.7\% for RLinear, 10.0\% for RMLP, 8.8\% for iTransformer, and 12.5\% for PatchTST, with corresponding MAE improvements of 3.3\%, 6.8\%, 4.7\%, and 8.6\% respectively.

\begin{figure}[h]
\centering
\includegraphics[width=0.99\linewidth]{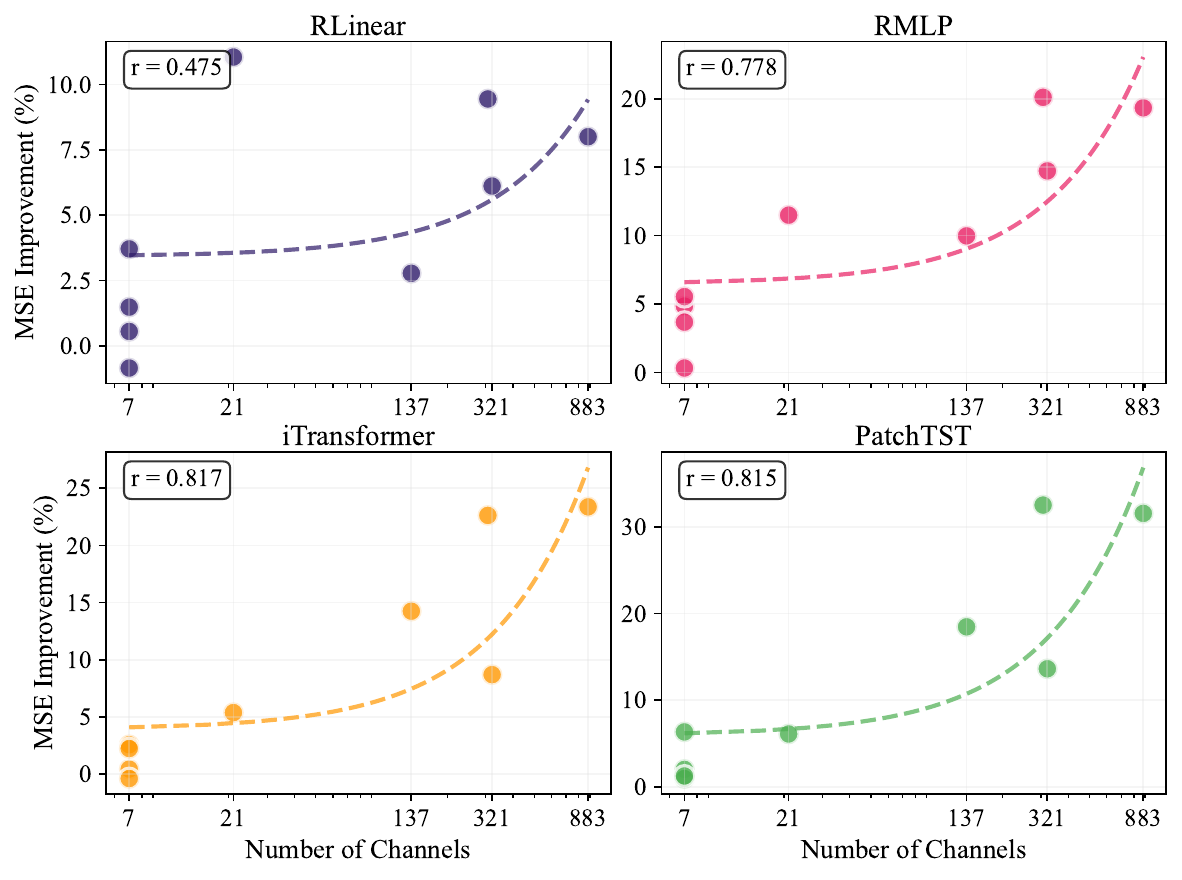}
\caption{Performance improvement vs. number of channels across four backbone models.}
\label{fig:channel_improvement}
\end{figure}

These improvements suggest effectiveness across diverse architectural paradigms: patch-based vs. variable-based embeddings, and channel-independent vs. channel-dependent models. Notably, even channel-dependent models like iTransformer benefit significantly, as they still rely on shared embedding layers that constrain channel-specific representation learning from the beginning. LightSAE addresses this bottleneck by enabling heterogeneous representations right from the embedding stage.

An important observation is the correlation between dataset characteristics and improvement magnitude. As visualized in Fig.~\ref{fig:channel_improvement}, performance improvements show a positive correlation with channel count across all backbone models (r=0.475-0.817). For low-dimensional IoT systems (ETT family with 7 channels), LightSAE achieves moderate improvements (1-6\%). In contrast, for large-scale IoT deployments with hundreds of channels (Electricity smart grids with 321 sensors, PEMS transportation systems with 307-883 sensors), performance gains can exceed 20\% in MSE. This observation suggests that as IoT systems become more complex with additional sensors, the benefits of modeling channel heterogeneity through our proposed approach become more pronounced, likely due to increased inter-channel diversity that shared embeddings cannot adequately capture.

\subsection{Ablation Studies}
\label{ssec:ablation}

To evaluate our core hypotheses and understand each component's contribution, we conduct a comprehensive ablation study evaluating: (i) the impact of SAE decomposition versus purely independent counterparts, and (ii) the effectiveness of different exploitation strategies for the structural observations within the SAE framework.

Table~\ref{tab:ablation_design} provides the mathematical formulations for each variant, while Table~\ref{tab:ablation_results} presents an ablation analysis using the iTransformer backbone. The Configuration columns show different methods organized by embedding framework: Shared (standard shared embedding), Ind (purely independent embedding), and SAE (our proposed framework). The LR and Pool columns indicate whether each method uses Low-Rank factorization and Shared Pool components respectively.

\begin{table}[h]
\centering
\caption{Mathematical formulations for different ablation study variants.}
\label{tab:ablation_design}
\resizebox{\linewidth}{!}{%
\renewcommand{\arraystretch}{1.3}
\begin{tabular}{l|c|c|c|l}
\toprule
Method & Frame. & LR & Pool & Mathematical Formulation \\
\midrule
Baseline & Shared & $\times$ & $\times$ & $\bm{e}_i = \bm{X}_i \bm{W}_{\mathrm{sh}}$ \\
\midrule
Ind-Full & Ind & $\times$ & $\times$ & $\bm{e}_i = \bm{X}_i \bm{W}_i$ \\
Ind-LR & Ind & $\checkmark$ & $\times$ & $\bm{e}_i = \bm{X}_i \bm{L}_i \bm{R}_i$ \\
Ind-Pool & Ind & $\times$ & $\checkmark$ & $\bm{e}_i = \bm{X}_i \sum_k g_{i,k} \bm{W}_k$ \\
LightSAE-Ind & Ind & $\checkmark$ & $\checkmark$ & $\bm{e}_i = \bm{X}_i (\sum_k g_{i,k} \bm{L}_k) \bm{R}_{\text{pool}}$ \\
\midrule
SAE-Full & SAE & $\times$ & $\times$ & $\bm{e}_i = \bm{X}_i (\bm{W}_{\mathrm{sh}} + \bm{W}_{c_i})$ \\
SAE-LR & SAE & $\checkmark$ & $\times$ & $\bm{e}_i = \bm{X}_i (\bm{W}_{\mathrm{sh}} + \bm{L}_i \bm{R}_i)$ \\
SAE-Pool & SAE & $\times$ & $\checkmark$ & $\bm{e}_i = \bm{X}_i (\bm{W}_{\mathrm{sh}} + \sum_k g_{i,k} \bm{W}_k)$ \\
LightSAE & SAE & $\checkmark$ & $\checkmark$ & $\bm{e}_i = \bm{X}_i (\bm{W}_{\mathrm{sh}} + (\sum_k g_{i,k} \bm{L}_k) \bm{R}_{\text{pool}})$ \\
\bottomrule
\end{tabular}%
}
\end{table}

\begin{table*}[!h]
\centering
\caption{Ablation study results with framework-based analysis using iTransformer backbone. The lookback window is fixed at $L=96$ and results are averaged across different prediction horizons $H \in \{96, 192, 336, 720\}$. We compare different embedding frameworks (Shared, Ind, SAE) and technical components (LR: Low-Rank, Pool: Shared Pool). Results show MSE, performance improvement (\%), parameter count (M), and parameter increase (\%).}
\label{tab:ablation_results}
\resizebox{0.9\textwidth}{!}{%
\renewcommand{\arraystretch}{1.1}
\begin{tabular}{l|c|c|c|rrrr|rrrr}
\toprule
\multicolumn{4}{c|}{Configuration} & \multicolumn{4}{c|}{Solar} & \multicolumn{4}{c}{PEMS04} \\
\cmidrule(lr){1-4} \cmidrule(lr){5-8} \cmidrule(l){9-12}
Method & Frame. & LR & Pool & MSE & $\Delta$\% & Params & $\Delta$\% & MSE & $\Delta$\% & Params & $\Delta$\% \\
\midrule
Baseline & Shared & \texttimes & \texttimes & 0.233 & $+$0.0\% & 3.38M & $+$0.0\% & 0.198 & $+$0.0\% & 4.96M & $+$0.0\% \\
\midrule
Ind-Full & Ind & \texttimes & \texttimes & 0.224 & $+$3.8\% & 10.13M & $+$199.9\% & 0.178 & $+$10.2\% & 20.15M & $+$306.6\% \\
Ind-LR & Ind & \checkmark & \texttimes & 0.223 & $+$4.5\% & 5.48M & $+$62.2\% & 0.176 & $+$11.2\% & 9.73M & $+$96.3\% \\
Ind-Pool & Ind & \texttimes & \checkmark & 0.220 & $+$5.7\% & 3.82M & $+$13.1\% & 0.175 & $+$11.5\% & 5.40M & $+$9.0\% \\
LightSAE-Ind & Ind & \checkmark & \checkmark & 0.232 & $+$0.7\% & 3.44M & $+$1.7\% & 0.201 & $-$1.4\% & 5.10M & $+$3.0\% \\
\midrule
SAE-Full & SAE & \texttimes & \texttimes & 0.219 & $+$6.1\% & 10.18M & $+$201.3\% & 0.174 & $+$12.4\% & 20.20M & $+$307.6\% \\
SAE-LR & SAE & \checkmark & \texttimes & 0.213 & $+$8.7\% & 5.53M & $+$63.7\% & 0.162 & $+$18.2\% & 9.78M & $+$97.3\% \\
SAE-Pool & SAE & \texttimes & \checkmark & 0.211 & $+$9.5\% & 3.87M & $+$14.6\% & 0.165 & $+$16.5\% & 5.45M & $+$10.0\% \\
\textbf{LightSAE} & SAE & \checkmark & \checkmark & \textbf{0.200} & \textbf{$+$14.4\%} & \textbf{3.49M} & \textbf{$+$3.2\%} & \textbf{0.153} & \textbf{$+$22.8\%} & \textbf{5.15M} & \textbf{$+$4.0\%} \\
\bottomrule
\end{tabular}
}
\end{table*}

\subsubsection{Effectiveness of SAE Decomposition}

This section validates the effectiveness of our proposed SAE decomposition framework in improving forecasting performance compared to alternative embedding strategies. We compare SAE-based methods against their Independent (Ind) framework counterparts to quantify the performance benefits of the shared-auxiliary decomposition.

The quantitative results in Table~\ref{tab:ablation_results} show the effectiveness of the SAE framework. Across all method pairs, SAE-based approaches generally outperform their Ind counterparts. For instance, comparing SAE-Full vs Ind-Full on PEMS04, the SAE framework achieves an MSE of 0.174 compared to 0.178 for the Ind framework, representing a 2.2\% improvement. This performance advantage suggests that even when both approaches have equivalent representational capacity, the SAE decomposition may create a more favorable optimization landscape by decoupling the learning of common and channel-specific patterns.

This optimization advantage becomes even more pronounced when examining the performance of LightSAE-Ind and LightSAE. Under the SAE framework, LightSAE achieves strong performance (0.153 MSE on PEMS04, a 22.8\% improvement over baseline). However, when implemented under the Ind framework (LightSAE-Ind), performance drops to 0.201 MSE, falling far short of the SAE version. \rev{This performance drop occurs because the design rationale of LightSAE is violated in the Ind setting. Specifically, our structural analysis shows that low-rank and clustering patterns are observed in the auxiliary weights of SAE, but are weaker in the Ind framework.
Therefore, applying these mechanisms without the shared component of SAE leads to suboptimal results for two key reasons. For Ind-LR, imposing a low-rank constraint on a weight matrix that does not empirically exhibit strong low-rankness (Fig.~\ref{fig:svd_decay_wc}) creates an information bottleneck, as it must compress both high-rank common patterns and low-rank specific details into a single, constrained transformation. Similarly, for Ind-Pool, the component sharing mechanism is less effective because, as shown in Fig.~\ref{fig:wc_clustering}, the independent weights lack the clear clustering structure needed for efficient sharing. In essence, the SAE framework is crucial because it creates the conditions under which these structural properties emerge, allowing them to be effectively exploited.}

\subsubsection{Effectiveness and Efficiency of LightSAE Components Design}

Having established that structural patterns are more apparent under SAE decomposition, we now dissect how LightSAE operationalizes these observations for improved performance and efficiency. The results in the lower part of Table~\ref{tab:ablation_results} illustrate this progression within the SAE framework. The SAE-Full model suggests that modeling heterogeneity is beneficial (MSE drops from 0.198 to 0.174 on PEMS04), but at an unsustainable parameter cost (+307.6\%).

Fortunately, our component-wise ablations within the SAE framework show how each observed characteristic translates into an effective inductive bias. The low-rank structure serves as a compression prior in SAE-LR, where enforcing a low-rank constraint on auxiliary weights operationalizes our structural observation. This acts as a regularizer, encouraging the model to capture only the most significant channel-specific deviations. The result is not just substantial parameter reduction (to +97.3\%), but also improved performance (MSE 0.162 on PEMS04), suggesting that the low-rank structure is an important characteristic. Similarly, the clustering structure functions as a sharing prior in SAE-Pool, which operationalizes our clustering observation by enabling channels to share a common pool of auxiliary components. This leverages the insight that channels naturally form groups with similar representational needs, allowing for both parameter reuse among similar channels and the selection of distinct components for dissimilar ones. SAE-Pool achieves competitive performance (MSE 0.165 on PEMS04) with a modest parameter budget (+10.0\%), confirming the effectiveness of this structural exploitation.

Finally, LightSAE synergistically combines both structural priors by using a shared pool of low-rank components, simultaneously applying compression and sharing mechanisms. This combination yields the best accuracy (0.153 MSE, a 22.8\% improvement over baseline) with minimal parameter overhead (+4.0\%). This demonstrates that the low-rank and clustering observations are complementary, and that designing an architecture that explicitly respects both achieves a better trade-off between expressiveness and efficiency.

\subsubsection{Analysis of Application Point}

A design choice of LightSAE is its application as an input embedding module. To validate this choice, we compare its performance when applied at different architectural positions. Since LightSAE targets linear transformations, it can be applied to replace any linear layer in the general MTSF pipeline shown in Fig.~\ref{fig:framework}(a). Specifically, we examine three configurations: (1) embedding layer only, (2) projection head (output layer) only, and (3) both layers. This analysis helps determine the optimal placement strategy for LightSAE within the model architecture.

\begin{figure}[h]
\centering
\includegraphics[width=0.85\linewidth]{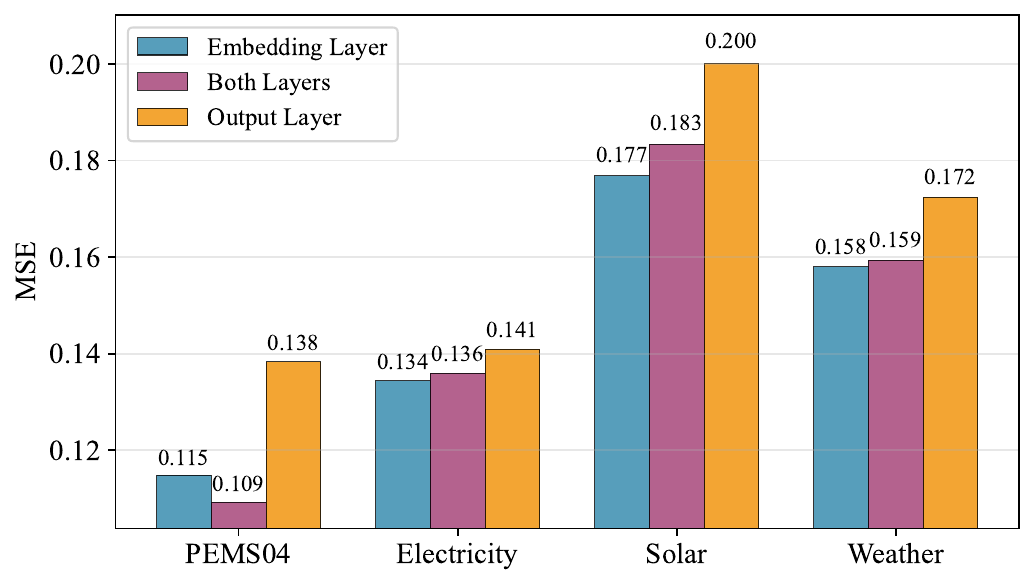}
\caption{Performance comparison of LightSAE applied at different positions in the model architecture (input length 96, prediction length 96).}
\label{fig:position_analysis}
\end{figure}

As shown in Fig.~\ref{fig:position_analysis}, the embedding layer approach achieves the best performance on three out of four datasets, while the output layer approach generally performs worst across all datasets. This supports our hypothesis that heterogeneity-aware modeling at the embedding stage enables all subsequent layers to benefit from channel-specific representations. While applying LightSAE at both layers can yield marginal gains in some cases (e.g., PEMS04), the initial input-level application appears to be the primary driver of performance improvement. Given the added parameter cost, applying LightSAE solely at the embedding layer offers a compelling parameter-performance trade-off. In contrast, addressing heterogeneity only at the output layer may fail to inform intermediate feature learning, limiting the model's ability to learn discriminative channel-specific patterns.

\subsubsection{Comparison with Related Methods}
\label{sssec:comparison_related_methods}

To further validate the effectiveness of our approach, we compare LightSAE against recent methods that can also model channel heterogeneity. These methods represent different approaches to modeling channel-specific patterns:

\textbf{C-LoRA} \cite{nie2024channel} applies low-rank adaptation to individual channels after token embedding to achieve channel-specific adaptation.

\textbf{MoLE} \cite{ni2024mixture} employs mixture-of-experts at the output layer with timestamp-dependent routing, training multiple linear experts where a router model adaptively weighs their outputs based on temporal periodicity.

\textbf{VE} \cite{wang2024ve} uses variate-dependent experts at the output layer, creating channel-specific representations through variate-aware expert selection to model heterogeneous channel patterns.

\begin{table}[h]
\centering
\caption{Performance comparison with related methods. Results show MSE and MAE values with look-back window 96 and averaged over four prediction horizons \{96, 192, 336, 720\}. Best results are highlighted in bold.}
\label{tab:comparison_related_methods}
\resizebox{0.5\textwidth}{!}{%
\renewcommand{\arraystretch}{1.15}
\begin{tabular}{l|l|cc|cc|cc|cc|cc}
\toprule
\multicolumn{2}{c|}{Methods} & \multicolumn{2}{c|}{Baseline} & \multicolumn{2}{c|}{+ C-LoRA} & \multicolumn{2}{c|}{+ MoLE} & \multicolumn{2}{c|}{+ VE} & \multicolumn{2}{c}{\textbf{+ LightSAE}} \\
\cmidrule(lr){3-4}\cmidrule(lr){5-6}\cmidrule(lr){7-8}\cmidrule(lr){9-10}\cmidrule(lr){11-12}
\multicolumn{2}{c|}{Metric} & MSE & MAE & MSE & MAE & MSE & MAE & MSE & MAE & MSE & MAE \\
\midrule
\multirow{5}{*}{\rotatebox[origin=c]{90}{RMLP}}
& Weather & 0.272 & 0.291 & 0.247 & 0.276 & 0.254 & 0.282 & 0.251 & 0.276 & \textbf{0.244} & \textbf{0.273} \\
& Solar & 0.261 & 0.313 & 0.245 & 0.291 & 0.246 & 0.291 & 0.241 & 0.289 & \textbf{0.234} & \textbf{0.281} \\
& Electricity & 0.213 & 0.295 & 0.198 & 0.287 & 0.201 & 0.290 & 0.195 & 0.285 & \textbf{0.182} & \textbf{0.271} \\
& PEMS04 & 0.290 & 0.376 & 0.240 & 0.341 & 0.245 & 0.348 & 0.243 & 0.338 & \textbf{0.232} & \textbf{0.329} \\
& PEMS07 & 0.345 & 0.400 & 0.290 & 0.353 & 0.295 & 0.355 & 0.302 & 0.361 & \textbf{0.279} & \textbf{0.346} \\
\midrule
\multirow{5}{*}{\rotatebox[origin=c]{90}{iTransformer}}
& Weather & 0.258 & 0.278 & 0.251 & 0.275 & 0.249 & 0.271 & 0.253 & 0.276 & \textbf{0.246} & \textbf{0.273} \\
& Solar & 0.233 & 0.262 & 0.211 & 0.261 & 0.214 & 0.273 & 0.209 & 0.260 & \textbf{0.200} & \textbf{0.260} \\
& Electricity & 0.178 & 0.270 & 0.168 & 0.265 & 0.171 & 0.268 & 0.168 & 0.266 & \textbf{0.162} & \textbf{0.258} \\
& PEMS04 & 0.198 & 0.299 & 0.165 & 0.267 & 0.169 & 0.274 & 0.168 & 0.273 & \textbf{0.153} & \textbf{0.257} \\
& PEMS07 & 0.194 & 0.286 & 0.160 & 0.262 & 0.165 & 0.269 & 0.163 & 0.269 & \textbf{0.149} & \textbf{0.237} \\
\bottomrule
\end{tabular}
}
\end{table}

The results in Table~\ref{tab:comparison_related_methods} provide empirical support for our design choices, demonstrating the importance of both architectural positioning and our SAE framework.

First, the consistent performance advantage of LightSAE over MoLE and VE suggests the importance of \textbf{architectural positioning}. By specializing only at the output layer, MoLE and VE are fundamentally limited, as they operate on feature representations that have already been homogenized by a shared input embedding. LightSAE's superior results suggest that to effectively model heterogeneity, intervention at the initial embedding stage is more effective for preserving and propagating channel-specific information throughout the model.

Second, the results highlight the important role of our \textbf{SAE decomposition} in providing a clear motivation for efficient heterogeneity modeling. Methods like MoLE and VE lack this decomposition entirely; consequently, their use of expert-mixing is not guided by the low-rank and clustering priors that we observe in our SAE analysis. In contrast, C-LoRA does leverage low-rank adaptation but fails to exploit the clustering structure. This leads to an inefficient design where a distinct adapter is required for every channel, overlooking the potential for sharing components among similar channels. In contrast, LightSAE synergistically leverages the low-rank and clustering characteristics from the SAE decomposition, achieving parameter efficiency and high performance.

\subsection{Hyperparameter Sensitivity}
\label{ssec:hyper_sensitivity}

We first analyze the sensitivity of LightSAE to its two main hyperparameters: rank $r$ and pool size $K$.

\begin{figure}[ht]
\centering
\includegraphics[width=0.99\linewidth]{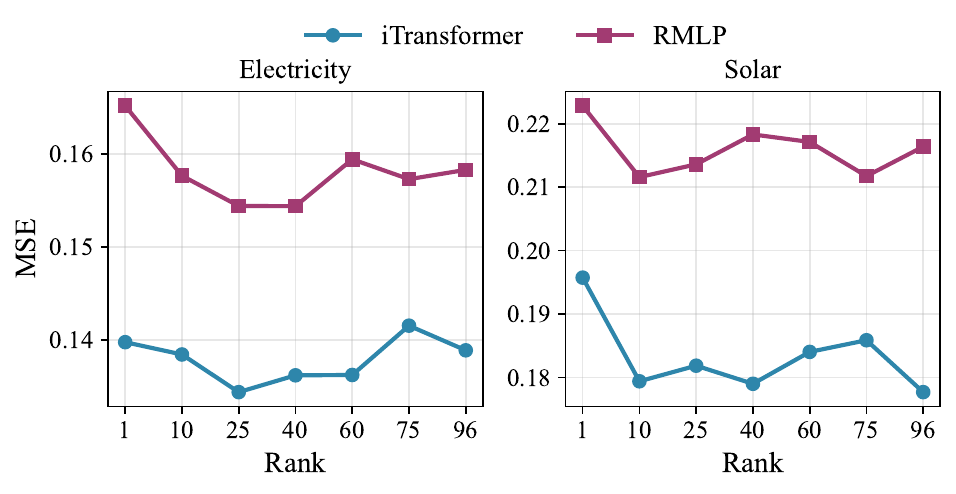}
\caption{Model performance under different rank values.}
\label{fig:rank_sensitivity}
\end{figure}

As shown in Fig.~\ref{fig:rank_sensitivity}, the model exhibits robust performance across different rank values. For iTransformer, performance generally improves as the rank increases from $r=1$, reaching an optimal range around $r=25$ to $r=40$, before slightly degrading at higher ranks. This suggests that a moderate rank is sufficient to capture essential channel-specific information without overfitting.

\begin{figure}[ht]
\centering
\includegraphics[width=0.99\linewidth]{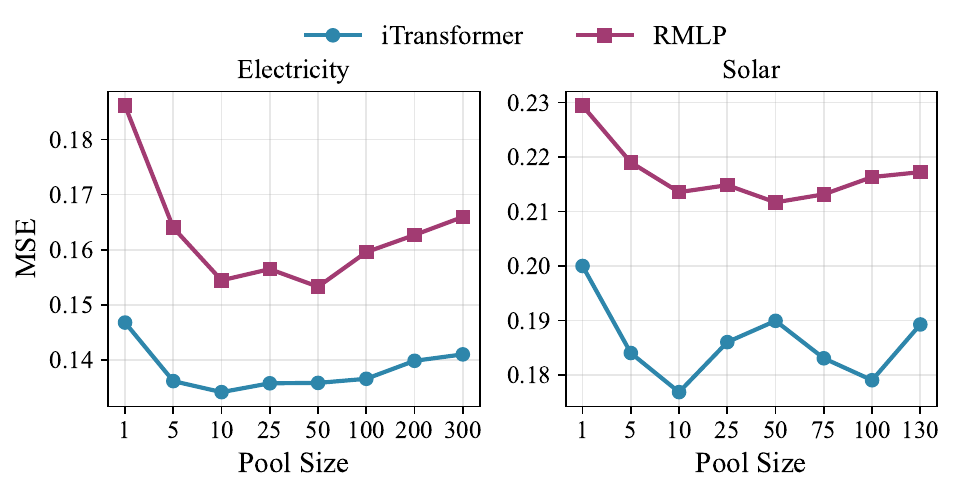}
\caption{Model performance under different pool size values.}
\label{fig:num_experts_sensitivity}
\end{figure}

Similarly, Fig.~\ref{fig:num_experts_sensitivity} illustrates the model's robustness to the pool size $K$. Both models show a trend where performance improves significantly when moving from $K=1$ to $K=10$, indicating the benefit of a diverse component pool. Beyond this point, performance stabilizes, with $K=10$ representing a good trade-off that balances expressiveness and parameter efficiency. The overall stability across a wide range of both $r$ and $K$ values suggests the robustness of our LightSAE module.

\begin{figure}[ht]
\centering
\includegraphics[width=0.99\linewidth]{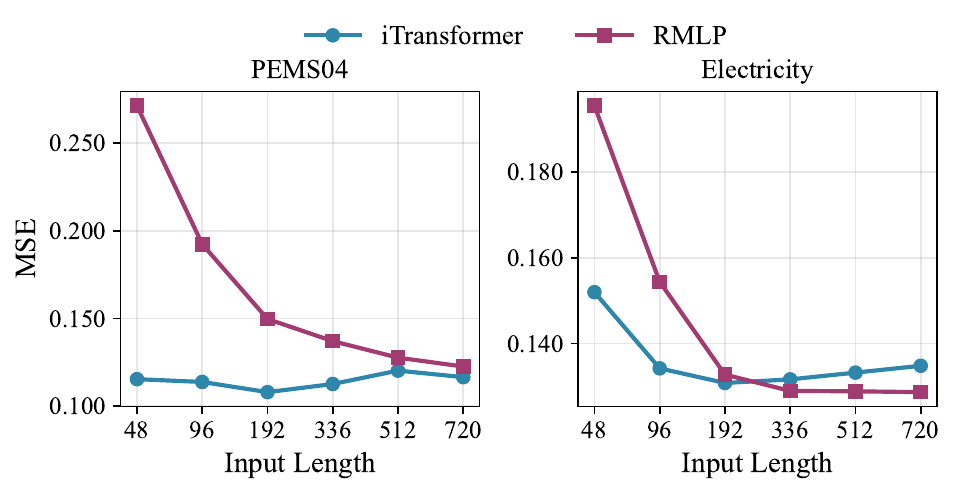}
\caption{Performance across different input sequence lengths.}
\label{fig:input_length}
\end{figure}

We further evaluate the robustness of LightSAE to variations in input sequence length. As shown in Fig.~\ref{fig:input_length}, both backbone models generally benefit from increased input lengths, with RMLP exhibiting more pronounced improvements. For iTransformer, performance improves with longer sequences but shows slight degradation at very long lengths. Overall, these results indicate that using LightSAE still enables the model to benefit from increased input sequence length.

\subsection{Visualization Analysis}
\label{ssec:visualization_analysis}
We conduct visualization analyses of LightSAE using the RMLP backbone on the Electricity dataset.

\subsubsection{Gating Weights Analysis}

\begin{figure*}[ht]
    \centering
    \includegraphics[width=0.85\linewidth]{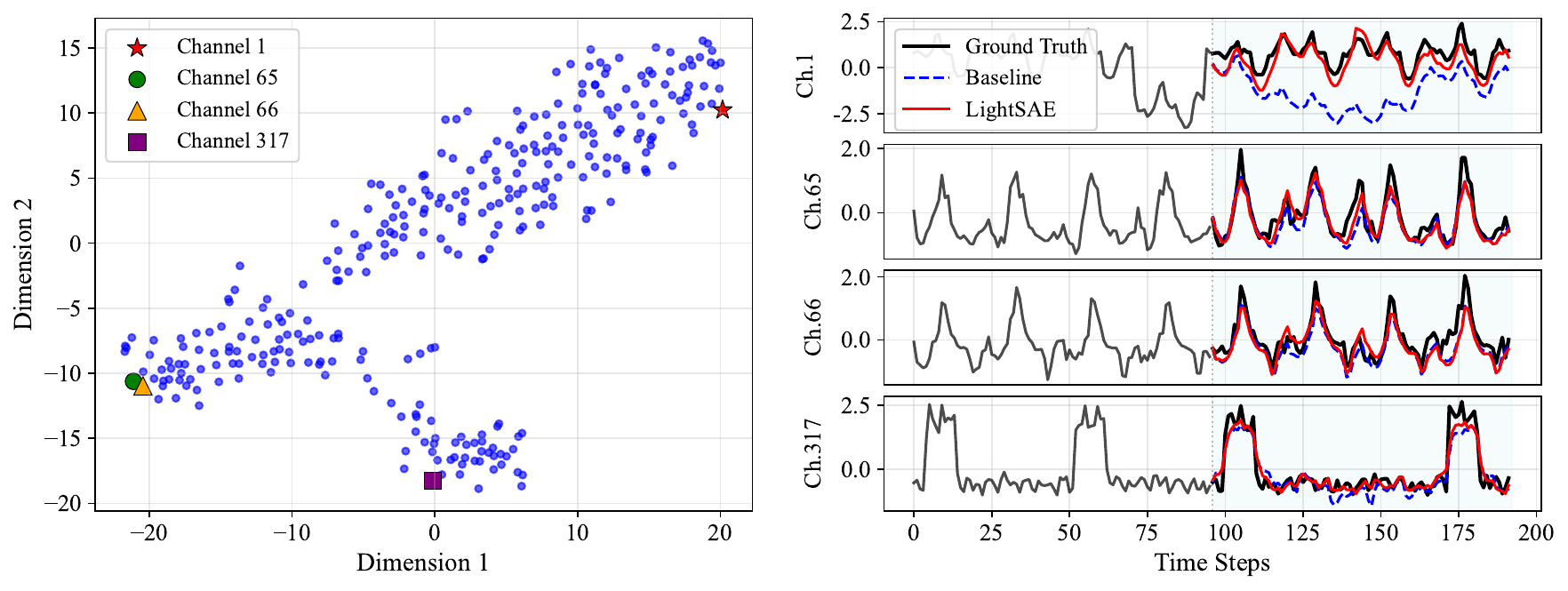}
    \caption{Visualization of learned gating weights using t-SNE (left) and prediction performance on representative channels (right). }
    \label{fig:gate_visualization}
\end{figure*}

Fig.~\ref{fig:gate_visualization} shows a t-SNE \cite{maaten2008visualizing} projection of the $K$-dimensional gating weight vectors $\{g_{i,k}\}$ for all channels. Channels with proximate positions in the embedding space exhibit similar temporal dynamics (e.g., channels 65 and 66 show nearly identical oscillatory patterns), while distant channels display substantial differences (e.g., channel 317 exhibits distinct spike patterns). This spatial-temporal correspondence suggests that LightSAE's gating mechanism captures meaningful channel relationships based on temporal characteristics. The right panel shows prediction performance on representative channels, where the baseline struggles while LightSAE closely follows the ground truth, particularly evident in channels 1 and 317.

\subsubsection{Auxiliary Component Analysis}

\begin{figure*}[ht]
    \centering
    \includegraphics[width=0.85\linewidth]{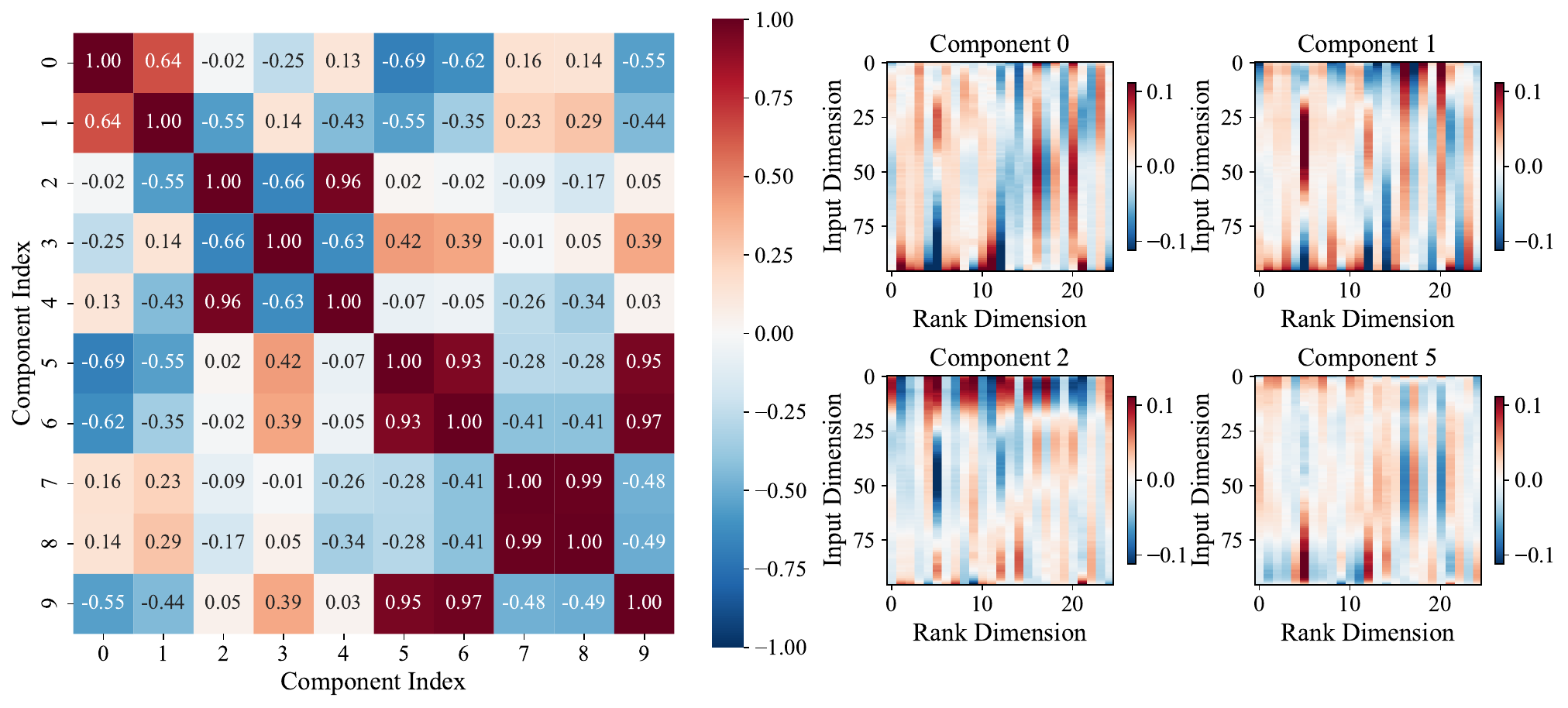}
    \caption{Analysis of auxiliary component pool in LightSAE. (Left) Pairwise cosine similarity heatmap. \rev{(Right) Visualization of the left-side low-rank matrices, $\bm{L}_k \in \mathbb{R}^{L \times r}$. For each component heatmap, the Y-axis represents input sequence length ($L$) and the X-axis represents the rank of the low-rank matrices ($r$).}}
    \label{fig:auxiliary_component_analysis}
\end{figure*}

\rev{Fig.~\ref{fig:auxiliary_component_analysis} provides a visual analysis of the learned component pool. The heatmap on the left shows the cosine similarities between component matrices $\bm{L}_k \in \mathbb{R}^{L \times r}$. A key observation is the diversity within the pool, evidenced by the prevalence of component pairs with low similarity or near-orthogonality. This diversity suggests the pool has learned a varied set of components that can function as a basis. Such a structure is beneficial as it allows channels with similar characteristics to select and reuse a common subset of components, while enabling channels with unique patterns to compose distinct transformations from other, dissimilar components, thus allowing distinct transformations for different channel clusters while maintaining parameter efficiency. The visualizations on the right offer a qualitative confirmation, showing that components with high similarity (e.g., 0.64 for components 0 and 1) share visual patterns, whereas orthogonal ones (e.g., 0.02 for components 2 and 5) exhibit no discernible visual correlation. At the same time, the heatmap also shows components of high similarity (e.g., over 0.9 for components 5, 6, and 9), indicating that some redundancy still exists within the learned pool. While these highly similar, but not identical components may capture subtle variations of a common pattern, this observation also points to a potential avenue for future improvement, such as incorporating an orthogonality constraint during training to further encourage diversity and reduce redundancy.}

\section{Limitations}
\label{sec:limitations}

While LightSAE demonstrates strong empirical performance, we acknowledge several limitations that suggest avenues for future work. \rev{First, our design is empirically motivated by the observed low-rank and clustering characteristics. While we provide conceptual intuition, a formal theoretical explanation for why and under what conditions these structures emerge remains an open question. Consequently, its effectiveness may be reduced on datasets where such structural patterns are much less pronounced. Furthermore, our evaluation is focused on IoT-related benchmarks; future work could explore its generalizability to other time series domains such as finance or healthcare. Second, while our extensive experiments across nine representative datasets did not show evidence of the gating mechanism collapsing to a trivial solution, we acknowledge that such a risk could theoretically exist. However, our current framework does not include explicit regularization mechanisms (e.g., load balancing losses~\cite{JMLR:v23:21-0998}) to formally prevent collapsing. Incorporating such regularization to enhance robustness is valuable, and we leave it for future exploration.} Finally, as our approach is based on a linear additive decomposition, exploring non-linear extensions could further improve the modeling of complex heterogeneity patterns. We leave these promising directions for future exploration.
\section{Conclusion}
\label{sec:conclusion}

This work challenges the prevailing one-size-fits-all embedding paradigm in MTSF by addressing the issue of channel heterogeneity, which is particularly pronounced in IoT systems with diverse sensor networks. We introduced the Shared-Auxiliary Embedding (SAE) framework, which decomposes representations into shared and auxiliary components. Within this decomposition, we empirically observed that channel-specific information, when disentangled from common patterns, exhibits low-rank and clustering structures. This observation directly motivated the design of LightSAE, a module that operationalizes these characteristics through low-rank factorization and a shared, gated component pool that promotes both parameter sharing and channel-specific specialization. Experiments across 9 IoT-related datasets and 4 backbone architectures support our approach, showing that LightSAE can deliver substantial performance gains (e.g., up to 22.8\% MSE improvement with a minimal 4.0\% parameter increase). As a plug-and-play module, LightSAE offers a practical and effective solution for enhancing channel-specific representation learning in existing MTSF models. \rev{Given the effectiveness demonstrated across these diverse and representative datasets, we anticipate that the observed structural patterns and the proposed LightSAE framework will be beneficial for a broader range of IoT-related time series applications.}



\bibliography{IEEEabrv,ref.bib}

\begin{thebibliography}{10}
\providecommand{\url}[1]{#1}
\csname url@samestyle\endcsname
\providecommand{\newblock}{\relax}
\providecommand{\bibinfo}[2]{#2}
\providecommand{\BIBentrySTDinterwordspacing}{\spaceskip=0pt\relax}
\providecommand{\BIBentryALTinterwordstretchfactor}{4}
\providecommand{\BIBentryALTinterwordspacing}{\spaceskip=\fontdimen2\font plus
\BIBentryALTinterwordstretchfactor\fontdimen3\font minus \fontdimen4\font\relax}
\providecommand{\BIBforeignlanguage}[2]{{%
\expandafter\ifx\csname l@#1\endcsname\relax
\typeout{** WARNING: IEEEtran.bst: No hyphenation pattern has been}%
\typeout{** loaded for the language `#1'. Using the pattern for}%
\typeout{** the default language instead.}%
\else
\language=\csname l@#1\endcsname
\fi
#2}}
\providecommand{\BIBdecl}{\relax}
\BIBdecl

\bibitem{10980332}
L.~Shen, Y.~Wang, X.~Fan, Y.~Wei, and H.~Qiu, ``{Variable-Dynamic Multivariate Time-Series Forecasting for IoT Systems},'' \emph{IEEE Internet of Things Journal}, vol.~12, no.~14, pp. 28\,479--28\,492, 2025.

\bibitem{zhang2024sageformer}
Z.~Zhang, L.~Meng, and Y.~Gu, ``{SageFormer: Series-Aware Framework for Long-Term Multivariate Time-Series Forecasting},'' \emph{IEEE Internet of Things Journal}, vol.~11, no.~10, pp. 18\,435--18\,448, 2024.

\bibitem{qiu2024tfb}
X.~Qiu, J.~Hu, L.~Zhou, X.~Wu, J.~Du, B.~Zhang, C.~Guo, A.~Zhou, C.~S. Jensen, Z.~Sheng \emph{et~al.}, ``{TFB: Towards Comprehensive and Fair Benchmarking of Time Series Forecasting Methods},'' \emph{Proceedings of the VLDB Endowment}, vol.~17, no.~9, pp. 2363--2377, 2024.

\bibitem{kim2021reversible}
T.~Kim, J.~Kim, Y.~Tae, C.~Park, J.-H. Choi, and J.~Choo, ``{Reversible Instance Normalization for Accurate Time-series Forecasting against Distribution Shift},'' in \emph{International conference on learning representations}, 2021.

\bibitem{liu2022non}
Y.~Liu, H.~Wu, J.~Wang, and M.~Long, ``{Non-stationary Transformers: Exploring the Stationarity in Time Series Forecasting},'' \emph{Advances in neural information processing systems}, vol.~35, pp. 9881--9893, 2022.

\bibitem{electricityloaddiagrams20112014_321}
\BIBentryALTinterwordspacing
A.~Trindade, ``{ElectricityLoadDiagrams20112014},'' UCI Machine Learning Repository, 2015. [Online]. Available: \url{https://archive.ics.uci.edu/dataset/321/electricityloaddiagrams20112014}
\BIBentrySTDinterwordspacing

\bibitem{nie2022time}
Y.~Nie, N.~H. Nguyen, P.~Sinthong, and J.~Kalagnanam, ``{A Time Series is Worth 64 Words: Long-Term Forecasting with Transformers},'' in \emph{The Eleventh International Conference on Learning Representations}, 2023.

\bibitem{li2023revisiting}
Z.~Li, S.~Qi, Y.~Li, and Z.~Xu, ``{Revisiting Long-Term Time Series Forecasting: An Investigation on Linear Mapping},'' \emph{arXiv preprint arXiv:2305.10721}, 2023.

\bibitem{liu2024itransformer}
Y.~Liu, T.~Hu, H.~Zhang, H.~Wu, S.~Wang, L.~Ma, and M.~Long, ``{iTransformer: Inverted Transformers Are Effective for Time Series Forecasting},'' in \emph{The Twelfth International Conference on Learning Representations}, 2024.

\bibitem{DBLP:conf/icml/2021}
M.~Meila and T.~Zhang, Eds., \emph{Proceedings of the 38th International Conference on Machine Learning, {ICML} 2021, 18-24 July 2021, Virtual Event}, ser. Proceedings of Machine Learning Research, vol. 139.\hskip 1em plus 0.5em minus 0.4em\relax {PMLR}, 2021.

\bibitem{zhang2023meta}
Y.~Zhang, K.~Gong, K.~Zhang, H.~Li, Y.~Qiao, W.~Ouyang, and X.~Yue, ``Meta-transformer: A unified framework for multimodal learning,'' \emph{arXiv preprint arXiv:2307.10802}, 2023.

\bibitem{chen2025operational}
K.~Chen, T.~Han, F.~Ling, J.~Gong, L.~Bai, X.~Wang, J.-J. Luo, B.~Fei, W.~Zhang, X.~Chen \emph{et~al.}, ``The operational medium-range deterministic weather forecasting can be extended beyond a 10-day lead time,'' \emph{Communications Earth \& Environment}, vol.~6, no.~1, p. 518, 2025.

\bibitem{team2025gemma}
G.~Team, A.~Kamath, J.~Ferret, S.~Pathak, N.~Vieillard, R.~Merhej, S.~Perrin, T.~Matejovicova, A.~Ram{\'e}, M.~Rivi{\`e}re \emph{et~al.}, ``Gemma 3 technical report,'' \emph{arXiv preprint arXiv:2503.19786}, 2025.

\bibitem{wang2024deep}
Y.~Wang, H.~Wu, J.~Dong, Y.~Liu, M.~Long, and J.~Wang, ``{Deep Time Series Models: A Comprehensive Survey and Benchmark},'' \emph{arXiv preprint arXiv:2407.13278}, 2024.

\bibitem{zeng2023transformers}
A.~Zeng, M.~Chen, L.~Zhang, and Q.~Xu, ``{Are Transformers Effective for Time Series Forecasting?}'' in \emph{Proceedings of the AAAI conference on artificial intelligence}, vol.~37, no.~9, 2023, pp. 11\,121--11\,128.

\bibitem{wu2023timesnet}
H.~Wu, T.~Hu, Y.~Liu, H.~Zhou, J.~Wang, and M.~Long, ``{TimesNet: Temporal 2D-Variation Modeling for General Time Series Analysis},'' in \emph{The Eleventh International Conference on Learning Representations}, 2023.

\bibitem{chen2023tsmixer}
S.-A. Chen, C.-L. Li, S.~O. Arik, N.~C. Yoder, and T.~Pfister, ``{TSM}ixer: An all-{MLP} architecture for time series forecast-ing,'' \emph{Transactions on Machine Learning Research}, 2023.

\bibitem{zhu2024puyun}
S.~Zhu, Y.~Chen, P.~Yu, X.~Qu, Y.~Zhou, Y.~Ma, Z.~Zhao, Y.~Liu, H.~Mi, and B.~Wang, ``Puyun: Medium-range global weather forecasting using large kernel attention convolutional networks,'' \emph{arXiv preprint arXiv:2409.02123}, 2024.

\bibitem{shao2022spatial}
Z.~Shao, Z.~Zhang, F.~Wang, W.~Wei, and Y.~Xu, ``{Spatial-Temporal Identity: A Simple yet Effective Baseline for Multivariate Time Series Forecasting},'' in \emph{Proceedings of the 31st ACM international conference on information \& knowledge management}, 2022, pp. 4454--4458.

\bibitem{liu2023spatio}
H.~Liu, Z.~Dong, R.~Jiang, J.~Deng, J.~Deng, Q.~Chen, and X.~Song, ``{Spatio-Temporal Adaptive Embedding Makes Vanilla Transformer SOTA for Traffic Forecasting},'' in \emph{Proceedings of the 32nd ACM international conference on information and knowledge management}, 2023, pp. 4125--4129.

\bibitem{xiao2024gaformer}
J.~Xiao, R.~Liu, and E.~L. Dyer, ``{GAFormer: Enhancing Timeseries Transformers through Group-Aware Embeddings},'' in \emph{The Twelfth International Conference on Learning Representations}, 2024.

\bibitem{nie2024channel}
T.~Nie, Y.~Mei, G.~Qin, J.~Sun, and W.~Ma, ``{Channel-Aware Low-Rank Adaptation in Time Series Forecasting},'' in \emph{Proceedings of the 33rd ACM International Conference on Information and Knowledge Management}, 2024, pp. 3959--3963.

\bibitem{ni2024mixture}
R.~Ni, Z.~Lin, S.~Wang, and G.~Fanti, ``{Mixture-of-Linear-Experts for Long-Term Time Series Forecasting},'' in \emph{International Conference on Artificial Intelligence and Statistics}.\hskip 1em plus 0.5em minus 0.4em\relax PMLR, 2024, pp. 4672--4680.

\bibitem{wang2024ve}
S.~Wang, Z.~Man, Z.~Cao, J.~Zheng, and Z.~Ge, ``{VE: Modeling Multivariate Time Series Correlation with Variate Embedding},'' in \emph{2024 International Conference on Advanced Mechatronic Systems (ICAMechS)}.\hskip 1em plus 0.5em minus 0.4em\relax IEEE, 2024, pp. 105--110.

\bibitem{wang2024timexer}
Y.~Wang, H.~Wu, J.~Dong, G.~Qin, H.~Zhang, Y.~Liu, Y.~Qiu, J.~Wang, and M.~Long, ``{TimeXer: Empowering Transformers for Time Series Forecasting with Exogenous Variables},'' in \emph{Proceedings of the 38th International Conference on Neural Information Processing Systems}, 2024, pp. 469--498.

\bibitem{zhang2023crossformer}
Y.~Zhang and J.~Yan, ``{Crossformer: Transformer Utilizing Cross-Dimension Dependency for Multivariate Time Series Forecasting},'' in \emph{The eleventh international conference on learning representations}, 2023.

\bibitem{liu2024unitst}
J.~Liu, C.~Liu, G.~Woo, Y.~Wang, B.~Hooi, C.~Xiong, and D.~Sahoo, ``{UniTST: Effectively Modeling Inter-Series and Intra-Series Dependencies for Multivariate Time Series Forecasting},'' in \emph{NeurIPS Workshop on Time Series in the Age of Large Models}, 2024.

\bibitem{DBLP:conf/acl/AghajanyanGZ20}
A.~Aghajanyan, S.~Gupta, and L.~Zettlemoyer, ``Intrinsic dimensionality explains the effectiveness of language model fine-tuning,'' in \emph{Proceedings of the 59th Annual Meeting of the Association for Computational Linguistics and the 11th International Joint Conference on Natural Language Processing, {ACL/IJCNLP} 2021, (Volume 1: Long Papers), Virtual Event, August 1-6, 2021}, 2021, pp. 7319--7328.

\bibitem{hu2022lora}
E.~J. Hu, yelong shen, P.~Wallis, Z.~Allen-Zhu, Y.~Li, S.~Wang, L.~Wang, and W.~Chen, ``{LoRA: Low-Rank Adaptation of Large Language Models},'' in \emph{International Conference on Learning Representations}, 2022.

\bibitem{gunasekar2017implicit}
S.~Gunasekar, B.~E. Woodworth, S.~Bhojanapalli, B.~Neyshabur, and N.~Srebro, ``Implicit regularization in matrix factorization,'' \emph{Advances in neural information processing systems}, vol.~30, 2017.

\bibitem{kim2025lora}
J.~Kim, J.~Kim, and E.~K. Ryu, ``Lora training provably converges to a low-rank global minimum or it fails loudly (but it probably won't fail),'' \emph{arXiv preprint arXiv:2502.09376}, 2025.

\bibitem{flury1984common}
B.~N. Flury, ``Common principal components in k groups,'' \emph{Journal of the American Statistical Association}, vol.~79, no. 388, pp. 892--898, 1984.

\bibitem{liu2012robust}
G.~Liu, Z.~Lin, S.~Yan, J.~Sun, Y.~Yu, and Y.~Ma, ``Robust recovery of subspace structures by low-rank representation,'' \emph{IEEE transactions on pattern analysis and machine intelligence}, vol.~35, no.~1, pp. 171--184, 2012.

\bibitem{gao2024units}
S.~Gao, T.~Koker, O.~Queen, T.~Hartvigsen, T.~Tsiligkaridis, and M.~Zitnik, ``{UNITS: A Unified Multi-task Time Series Model},'' \emph{Advances in Neural Information Processing Systems}, vol.~37, pp. 140\,589--140\,631, 2024.

\bibitem{zhou2021informer}
H.~Zhou, S.~Zhang, J.~Peng, S.~Zhang, J.~Li, H.~Xiong, and W.~Zhang, ``{Informer: Beyond Efficient Transformer for Long Sequence Time-Series Forecasting},'' in \emph{Proceedings of the AAAI conference on artificial intelligence}, vol.~35, no.~12, 2021, pp. 11\,106--11\,115.

\bibitem{maaten2008visualizing}
L.~v.~d. Maaten and G.~Hinton, ``{Visualizing Data using t-SNE},'' \emph{Journal of machine learning research}, vol.~9, no. Nov, pp. 2579--2605, 2008.

\bibitem{JMLR:v23:21-0998}
\BIBentryALTinterwordspacing
W.~Fedus, B.~Zoph, and N.~Shazeer, ``Switch transformers: Scaling to trillion parameter models with simple and efficient sparsity,'' \emph{Journal of Machine Learning Research}, vol.~23, no. 120, pp. 1--39, 2022. [Online]. Available: \url{http://jmlr.org/papers/v23/21-0998.html}
\BIBentrySTDinterwordspacing

\end{thebibliography}
\bibliographystyle{IEEEtran}


\section{Biography}

\begin{IEEEbiography}[{\includegraphics[width=1in,height=1.25in,clip,keepaspectratio]{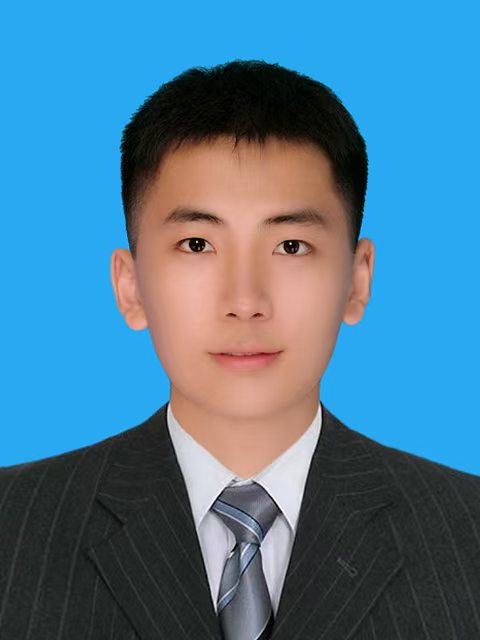}}]{Yi Ren}
was born in Hunan, China, in October 1998. He received the B.S. degree in electrical engineering from Tsinghua University, Beijing, China, in 2020. He is currently pursuing the Ph.D. degree in the Department of Electrical Engineering, Tsinghua University.
His research interests include artificial intelligence and data mining.
\end{IEEEbiography}

\begin{IEEEbiography}[{\includegraphics[width=1in,height=1.25in,clip,keepaspectratio]{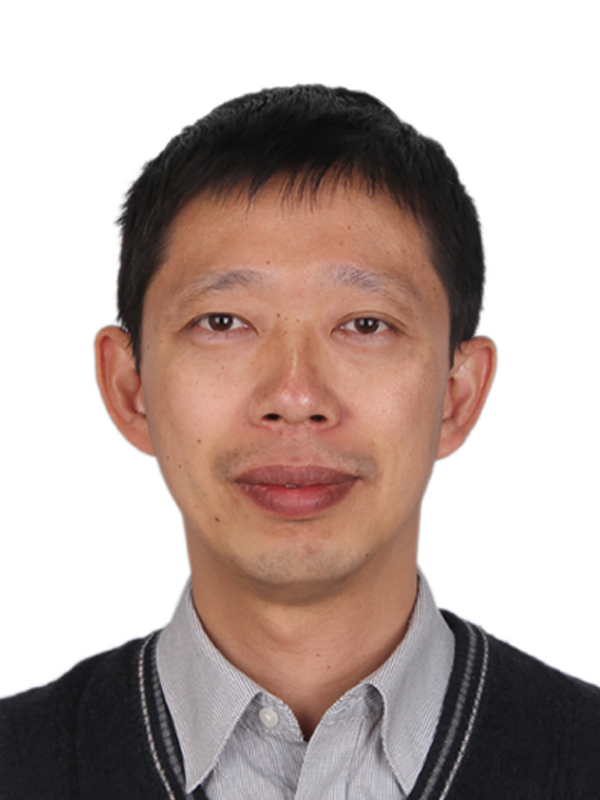}}]{Xinjie Yu}
(Senior Member, IEEE) was born in Guizhou, China, in February 1973. He received B.S. and Ph.D. degrees in electrical engineering from Tsinghua University, Beijing, China, in 1996 and 2001, respectively. He is currently a professor in the Department of Electrical Engineering, Tsinghua University.
His research interests include pulsed power supply, current sensors, and computational intelligence.
\end{IEEEbiography}

\vfill

\end{document}